\pdfoutput=1

\documentclass[11pt]{article}
\usepackage{enumitem}
\usepackage[preprint]{acl}
\usepackage{booktabs} 
\usepackage{times}
\usepackage{latexsym}

\usepackage[T1]{fontenc}

\usepackage[utf8]{inputenc}

\usepackage{microtype}

\usepackage{inconsolata}

\usepackage{graphicx}

\usepackage{colortbl}
\usepackage{pifont}
\usepackage{amssymb}
\usepackage{amsmath}
\usepackage{tcolorbox}
\usepackage{makecell} 

\usepackage{subfigure}
\definecolor{myblue}{rgb}{0.0,0.45,0.8}
\definecolor{myred}{rgb}{0.84,0.0,0.0}

\newcommand{\gaincell}[1]{%
  \ifdim #1 pt > 0pt
    \cellcolor{myblue!\numexpr min(100, abs(#1*2))\relax}{+#1\%}
  \else
    \cellcolor{myred!\numexpr min(100, abs(#1*2))\relax}{#1\%}
  \fi
}

\title{RAG in the Wild: On the (In)effectiveness of LLMs with \\ Mixture-of-Knowledge Retrieval Augmentation}

\author{Ran Xu$^\dagger$ \quad Yuchen Zhuang$^\ddagger$  \quad Yue Yu$^\ddagger$  \quad Haoyu Wang$^\#$  \quad Wenqi Shi$^\diamond$  \quad Carl Yang$^\dagger$  \\
  $^\dagger$ Emory University \quad 
  $^\ddagger$ Georgia Institute of Technology \\
  $^\#$ SUNY Albany \quad
  $^\diamond$ UT Southwestern Medical Center  \\
  \texttt{\{ran.xu, j.carlyang\}@emory.edu} \\}

\begin{document}
\maketitle
\begin{abstract}

Retrieval-augmented generation (RAG) enhances large language models (LLMs) by integrating external knowledge retrieved at inference time. 
While RAG demonstrates strong performance on benchmarks largely derived from general-domain corpora like Wikipedia, its effectiveness under realistic, diverse retrieval scenarios remains underexplored.
We evaluated RAG systems using \textsc{MassiveDS}, a large-scale datastore with mixture of knowledge, and identified critical limitations: retrieval mainly benefits smaller models, rerankers add minimal value, and no single retrieval source consistently excels. Moreover, current LLMs struggle to route queries across heterogeneous knowledge sources. These findings highlight the need for adaptive retrieval strategies before deploying RAG in real-world settings. 
Our code and data can be found at \url{https://github.com/ritaranx/RAG_in_the_Wild}.
\end{abstract}

\section{Introduction}
Retrieval-Augmented Generation (RAG) serves as a useful strategy for adapting large language models (LLMs) to tasks that require long-tailed or domain-specific knowledge. By retrieving relevant information from external knowledge sources, RAG provides LLMs with contextual evidence at inference time, enhancing performance on knowledge-intensive NLP tasks including question answering \citep{lewis2020retrieval, izacard2023atlas},  fact verification~\citep{lin2024radit}, and reasoning~\citep{islam-etal-2024-open,li2025search}.

Despite the strong performance of RAG on a range of benchmarks, these evaluations are predominantly constructed from, or closely aligned with Wikipedia~\citep[\emph{inter alia}]{nq,joshi-etal-2017-triviaqa,fever,hotpotqa,petroni-etal-2021-kilt,mallen-etal-2023-trust}. Consequently, high accuracy in these settings is somehow not particularly surprising, as many queries are well-covered by the retrieval corpus. 
In contrast, retrieval in real-world applications is substantially more challenging. The target corpus may be noisy, domain-specific, or misaligned with the query distribution. Although recent efforts have sought to evaluate RAG systems in broader knowledge tasks~\citep{asai2024selfrag,shi-etal-2024-replug,huang2024raven,zhang2024raft}, the underlying corpora in these studies only include Wikipedia or similar sources in the general domain, thus limiting their generalizability. 
These observations highlight a pressing need to evaluate RAG systems under real-world retrieval conditions. 

Motivated by these considerations, our goal is to comprehensively examine how corpus composition and model scale affect the performance of RAG systems under a more realistic setting. 
To this end, we consider a \emph{mixture-of-knowledge} scenario using \textsc{MassiveDS}~\citep{shao2025scaling} -- a large-scale, multi-domain datastore that combines general web sources (e.g., CommonCrawl) with specialized domains (e.g., PubMed). 
We evaluate tasks spanning both general knowledge and domain-specific QA, where \emph{no prior information} about the underlying knowledge source is available for each question. 
From the results, we have several findings: 

\begin{itemize}[leftmargin=0.35cm]
    \item Under the mixture-of-knowledge setting, the benefits of \emph{retrieval} are largely confined to smaller language models only -- once the backbone model becomes sufficiently powerful, these gains diminish significantly, except for factuality-focused QA tasks. 
    \item These diminishing returns aren't solely due to retrieval quality; adding a reranker offers marginal improvements, suggesting deeper integration challenges between retrieval and generation. 
    \item  No single source consistently outperforms others, including no-retrieval baselines, emphasizing the need for \emph{adaptive retrieval}. Yet, current LLMs struggle to route queries effectively across heterogeneous sources.
\end{itemize}
We believe our evaluations provide new insights and motivate future research towards more adaptive and robust retrieval-augmented generation systems under realistic, multi-domain conditions.

\begin{figure*}[t]
	\centering
    \vspace{-1ex}
	\subfigure[SimpleQA]{\includegraphics[width=0.315\linewidth]{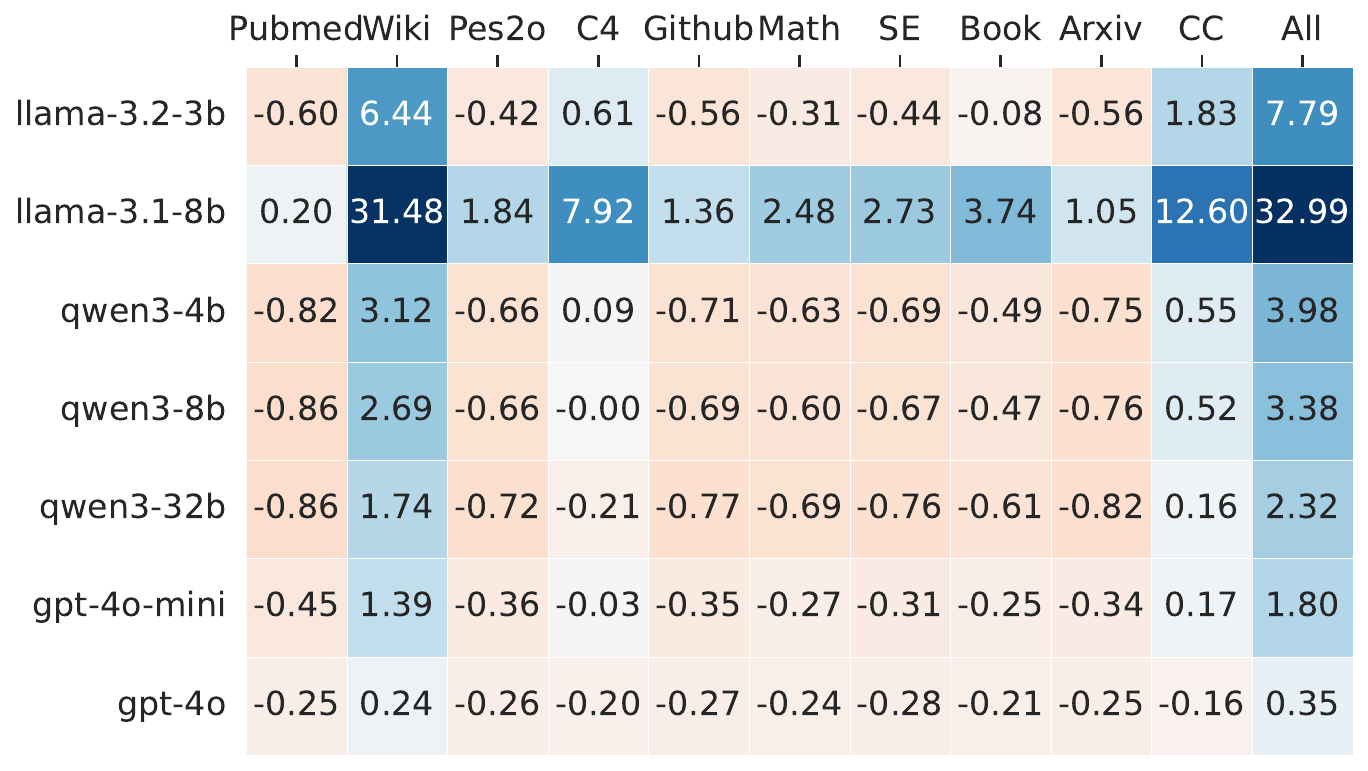}
		\label{fig:SimpleQA-no-wiki}
	} 
    \subfigure[SciQ]{\includegraphics[width=0.315\linewidth]{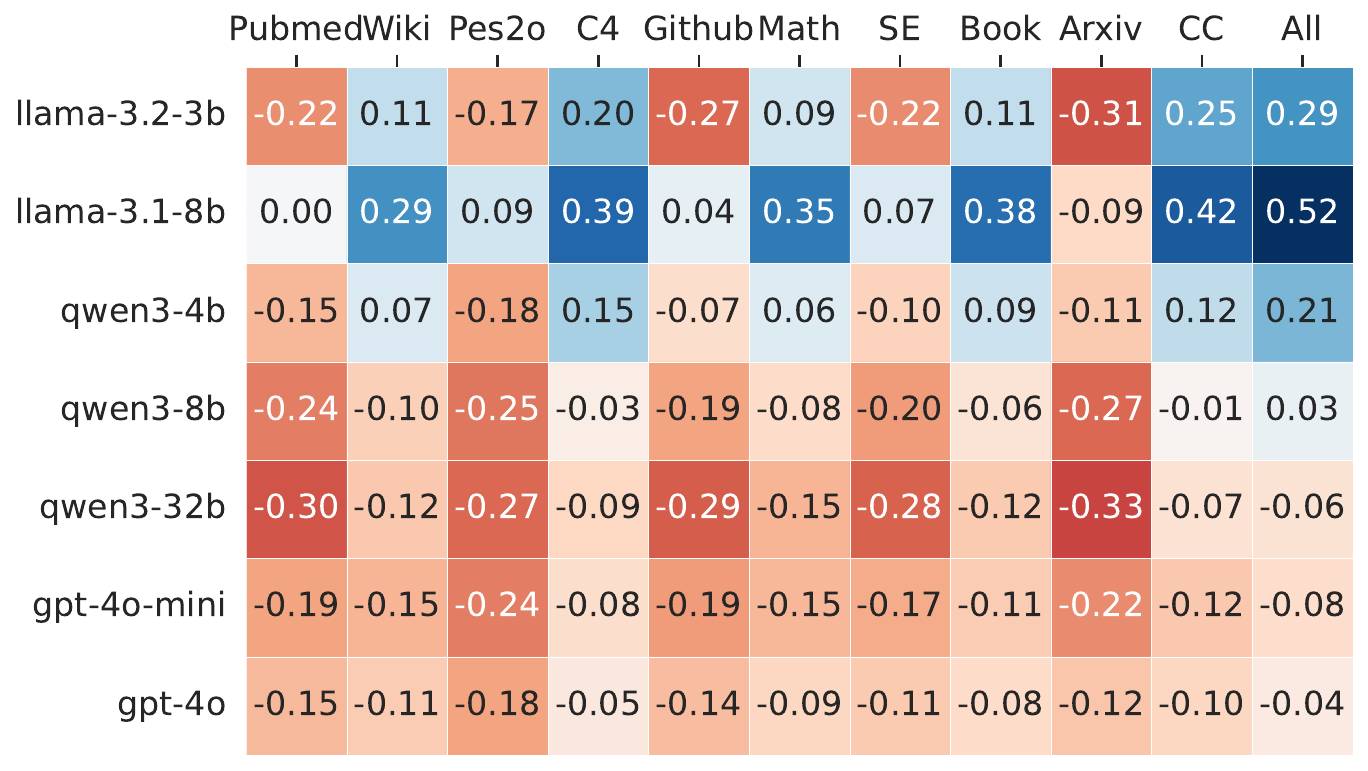}
		\label{fig:Science}
	}
    \subfigure[ARC-C]{
\includegraphics[width=0.315\linewidth]{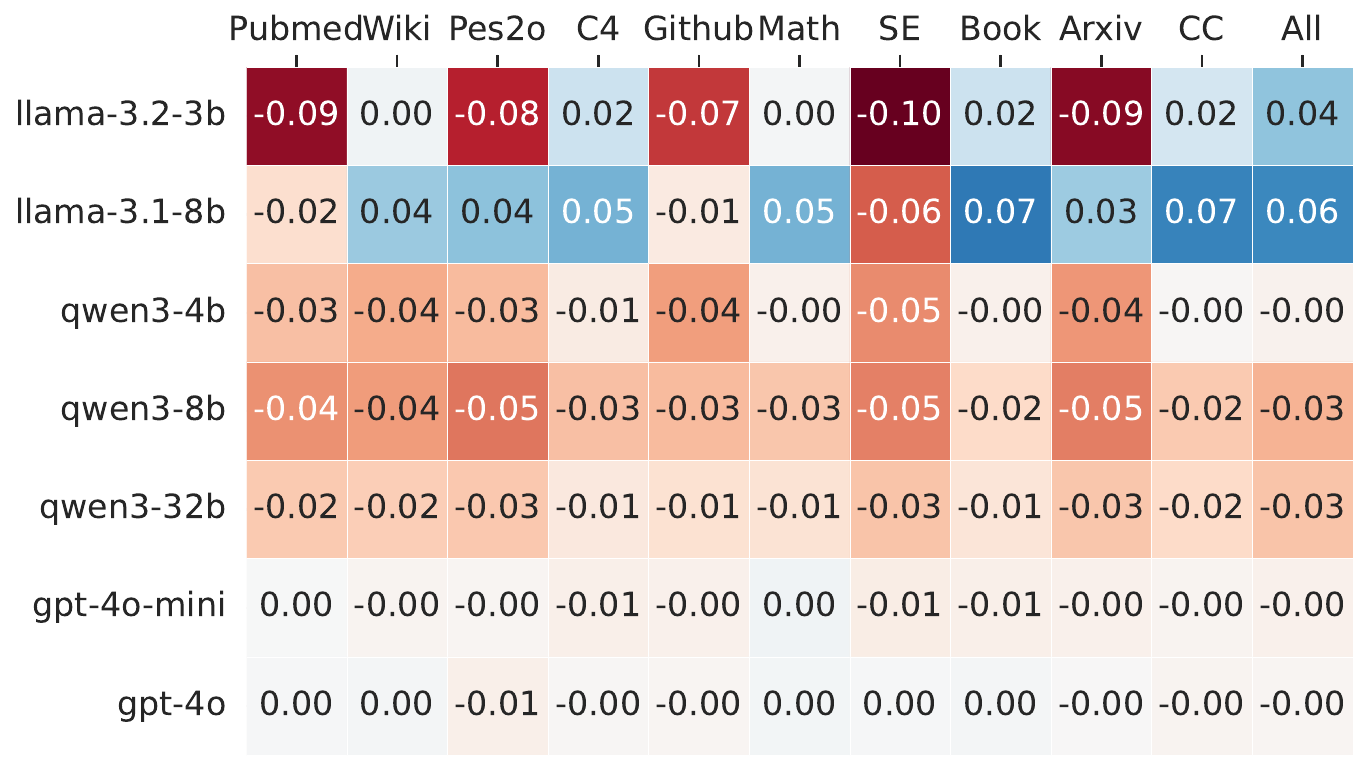}
		\label{fig:Science-arcc}
	}
    \subfigure[MMLU]{
\includegraphics[width=0.315\linewidth]{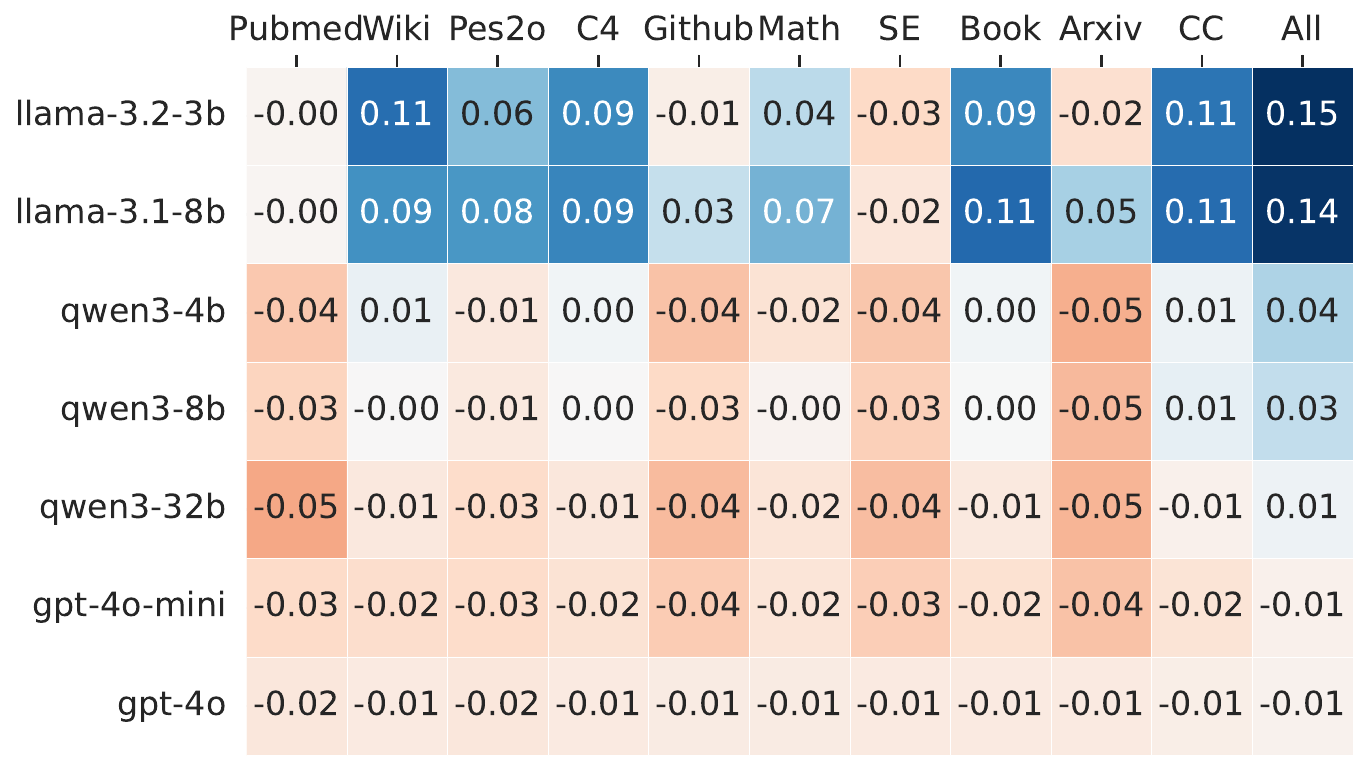}
		\label{fig:mmlu}
	} 
    	\subfigure[MMLU-Pro]{
		\includegraphics[width=0.315\linewidth]{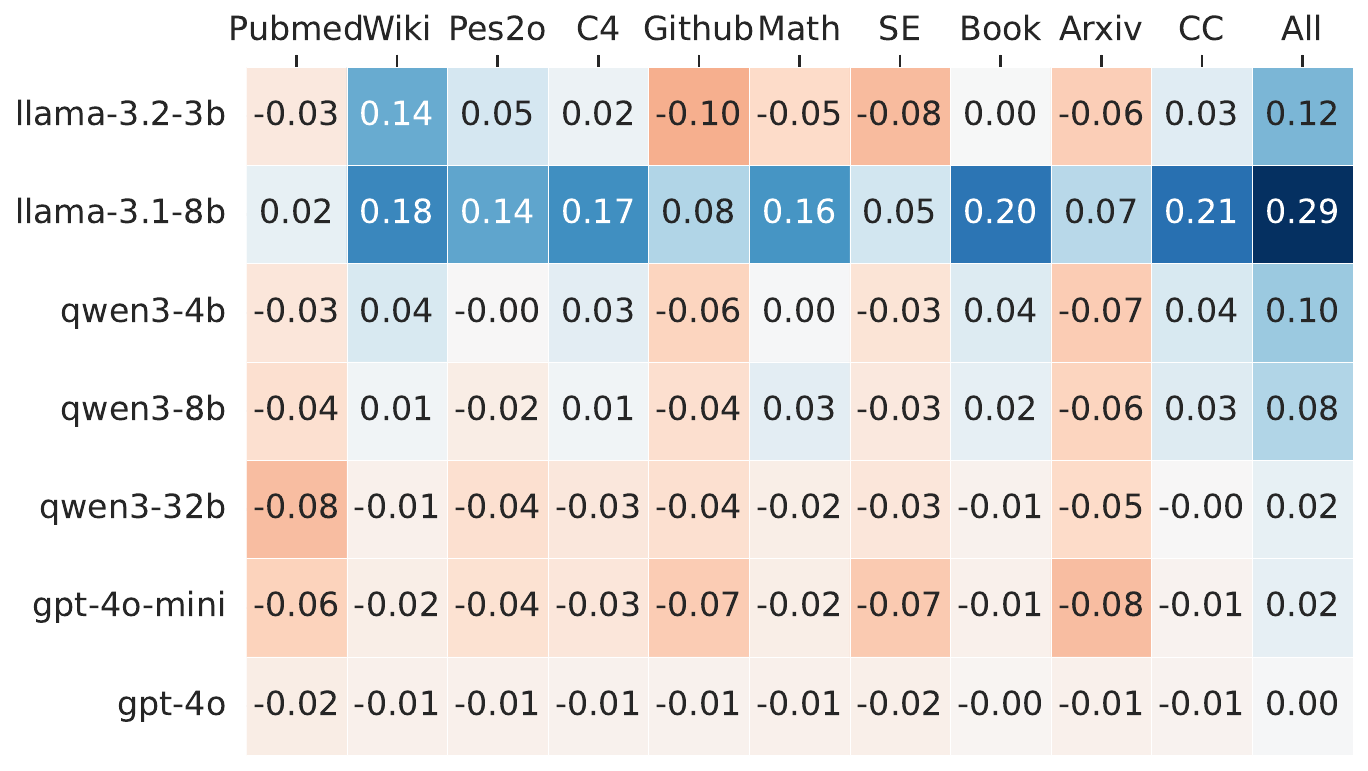}
		\label{fig:mmlu-pro}
	} 
    	\subfigure[CSBench]{
		\includegraphics[width=0.315\linewidth]{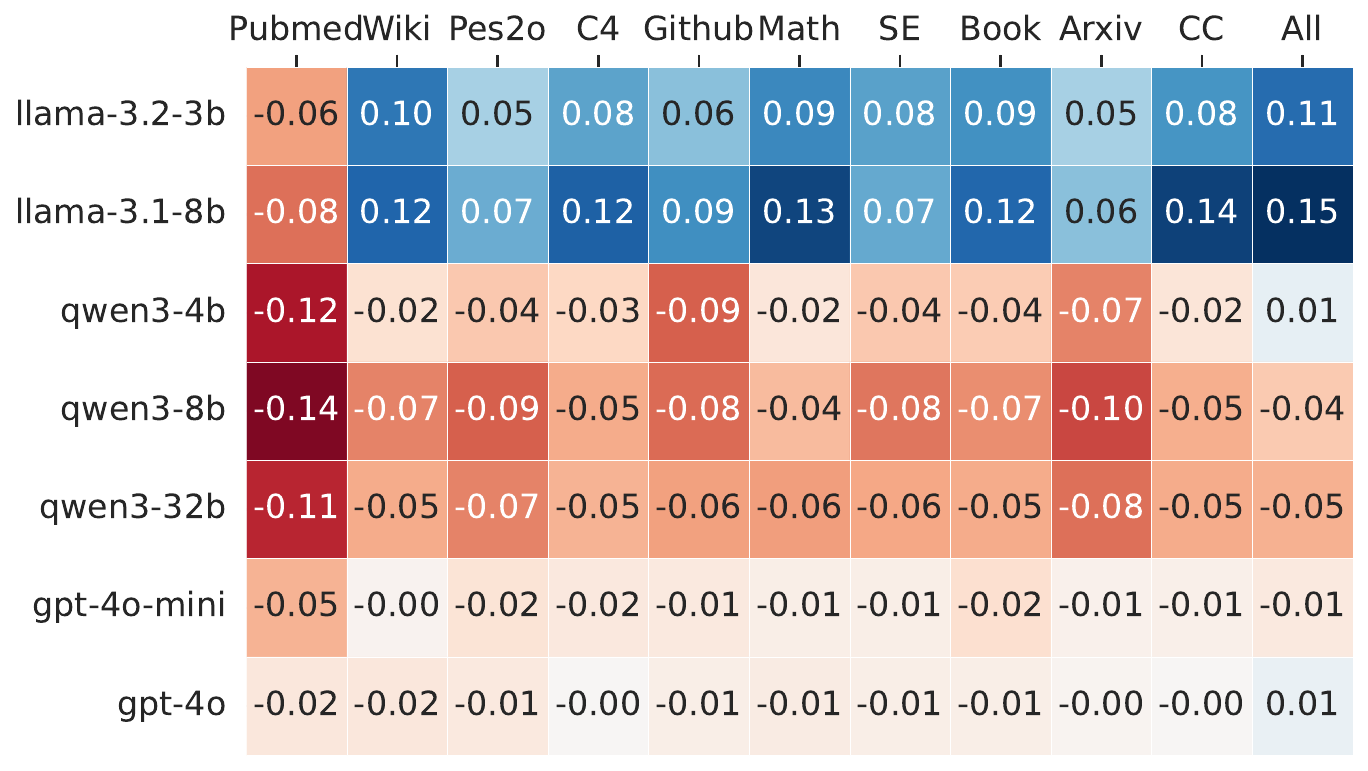}
		\label{fig:csbench}
	} 
     \vspace{-1ex}
	\caption{The relevance performance of different LLMs compared to non-retrieval baselines on six datasets. `All' means the corpus from all domains is considered for retrieval.\vspace{-0.5ex}
}
\label{fig:main}
\end{figure*}

\section{Experiments}
\subsection{Experiment Setup}
\noindent \textbf{Datasets.} 
To evaluate the accuracy of RAG systems across  domains, we identify two key desiderata: (1) \emph{Topic coverage}: the dataset should span a broad range of tasks that necessitate external knowledge for successful resolution; and (2) \emph{Corpus diversity}: questions should not be created solely from a single corpus (e.g., Wikipedia or PubMed). 

With this in mind, we select six datasets for our evaluation:
(1) \textbf{General knowledge-based QA}: we use MMLU~\cite{mmlu} and MMLU-Pro~\cite{mmlupro}, which contain questions spanning a wide array of subjects such as history, law, and medicine.
(2) \textbf{Scientific QA}: We evaluate on ARC Challenge~\cite{arcc} and SciQ~\cite{welbl2017crowdsourcing} for natural science, and CSBench~\cite{song2025csbench} for computer science, to test the model’s reasoning over domain-specific scientific knowledge. 
(3) \textbf{Factuality}: We adopt SimpleQA~\cite{wei2024measuring}, a benchmark collected from multiple web sources to test factual correctness. Note that we \emph{remove} those questions that were created using Wikipedia pages.

\noindent \textbf{Backbones.}
We consider three families of LLMs: (1) \textbf{Llama-series}: Llama-3.2-3B and Llama-3.1-8B \citep{grattafiori2024llama}; 
(2) \textbf{Qwen-series}: Qwen3-4B/8B/32B
\citep{yang2025qwen3};
(3) \textbf{GPT-series}: GPT-4o-mini/-4o~\citep{hurst2024gpt4o}. Note that we use \texttt{instruct} version in our experiments.

\noindent \textbf{Retrieval Corpora.}
We use \texttt{MassiveDS}, a massive datastore with 1.4T tokens as the knowledge source, where we consider PubMed, Wikipedia, Pes2o, C4, Github, Math, StackExchange (SE), Book, Arxiv and CommonCrawl (CC) as sources.

\noindent \textbf{Evaluation Metrics.}
Our evaluation covers multiple-choice tasks (MMLU, MMLU-Pro, ARC-C), short-form generation (SimpleQA, SciQ), and mixed formats (CSBench). We report accuracy for multiple-choice, exact match (EM) for generation, and follow \citet{song2025csbench} for CSBench. Retrieval effectiveness is measured by relative gain: 
\(\Delta(p_{s})=\frac{p_s-\rho}{\rho}\), 
where $p_s$ is the RAG performance using source $s$, and $\rho$ is performance of the LLM baseline without retrieval.

\noindent \textbf{Implementation Details.}
For Qwen-3, we use the non-reasoning model in our evaluation.
For retrieval, we use \emph{bge-base-en-v1.5} and \emph{bge-reranker-v2-m3}~\citep{chen2024bge} as the default retriever and reranker.
We evaluate \emph{zero-shot} performance with $k=5$ top passage.
For the reranking setting, we first retrieve $k'=30$ passages, then use the reranker to obtain $k=5$ passages with the highest relevance scores.  
Following \citet{shao2025scaling}, we apply document filtering and deduplication. 

\begin{table}[!t]
\centering
\footnotesize
\renewcommand{\arraystretch}{0.95}
\caption{Performance Gains of using Retrieval across different Domains in MMLU.}
\resizebox{\linewidth}{!}{
\begin{tabular}{lcccc}
\toprule
\textbf{Model} & \textbf{STEM} & \textbf{Social Sciences} & \textbf{Humanities} & \textbf{Others} \\
\midrule
Llama-3.2-3B        & 0.388 & 0.544 & 0.444 & 0.563 \\
w/ retrieval        & 0.477 & 0.644 & 0.478 & 0.649 \\
$\Delta$
    & \cellcolor{myblue!46}{+22.87\%}
    & \cellcolor{myblue!37}{+18.48\%}
    & \cellcolor{myblue!15}{+7.70\%}
    & \cellcolor{myblue!31}{+15.31\%} \\
\addlinespace
Llama-3.1-8B       & 0.472 & 0.598 & 0.492 & 0.578 \\
w/ retrieval        & 0.533 & 0.702 & 0.530 & 0.702 \\
$\Delta$
    & \cellcolor{myblue!26}{+12.93\%}
    & \cellcolor{myblue!35}{+17.35\%}
    & \cellcolor{myblue!16}{+7.85\%}
    & \cellcolor{myblue!43}{+21.39\%} \\
\addlinespace
Qwen-3-4B           & 0.653 & 0.761 & 0.576 & 0.707 \\
w/ retrieval        & 0.670 & 0.785 & 0.597 & 0.762 \\
$\Delta$
    & \cellcolor{myblue!5}{+2.57\%}
    & \cellcolor{myblue!6}{+3.16\%}
    & \cellcolor{myblue!7}{+3.57\%}
    & \cellcolor{myblue!16}{+7.91\%} \\
\addlinespace
Qwen-3-8B        & 0.677 & 0.790 & 0.598 & 0.750 \\
w/ retrieval        & 0.699 & 0.807 & 0.618 & 0.790 \\
$\Delta$
    & \cellcolor{myblue!8}{+3.19\%}
    & \cellcolor{myblue!6}{+2.13\%}
    & \cellcolor{myblue!8}{+3.24\%}
    & \cellcolor{myblue!15}{+5.29\%} \\
\addlinespace
Qwen3-32B         & 0.785 & 0.856 & 0.701 & 0.824 \\
w/ retrieval        & 0.803 & 0.851 & 0.705 & 0.827 \\
$\Delta$
    & \cellcolor{myblue!6}{+2.28\%}
    & \cellcolor{myred!10}{-0.50\%}
    & \cellcolor{myblue!2}{+0.61\%}
    & \cellcolor{myblue!2}{+0.41\%} \\
\addlinespace
GPT-4o-mini         & 0.686 & 0.853 & 0.671 & 0.816 \\
w/ retrieval        & 0.697 & 0.844 & 0.657 & 0.815 \\
$\Delta$
    & \cellcolor{myblue!4}{+1.53\%}
    & \cellcolor{myred!20.4}{-1.04\%}
    & \cellcolor{myred!40.2}{-2.12\%}
    & \cellcolor{myred!4.8}{-0.18\%} \\
\addlinespace
GPT-4o              & 0.773 & 0.900 & 0.806 & 0.867 \\
w/ retrieval        & 0.775 & 0.891 & 0.796 & 0.867 \\
$\Delta$
    & \cellcolor{myblue!0.4}{+0.34\%}
    & \cellcolor{myred!21.8}{-1.08\%}
    & \cellcolor{myred!24.8}{-1.28\%}
    & \cellcolor{myred!0.4}{-0.02\%} \\
\bottomrule
\end{tabular}
\label{tab:main}
}
\end{table}

\begin{figure*}[t]
	\centering
        \subfigure[LLaMA-3.2-3B]{
		\includegraphics[width=0.31\linewidth]{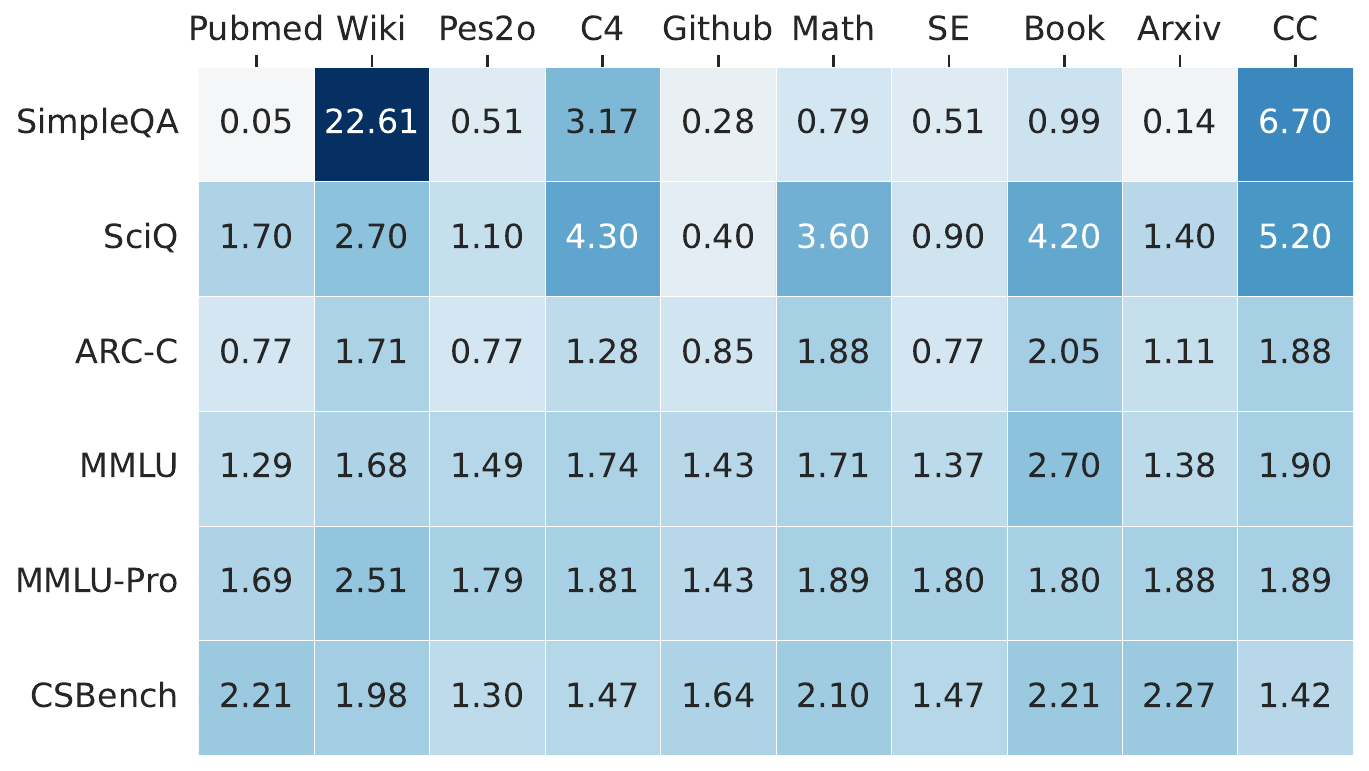}
		\label{fig:llama-3.2-3b}
	}
     \subfigure[LLaMA-3.1-8B]{
		\includegraphics[width=0.31\linewidth]{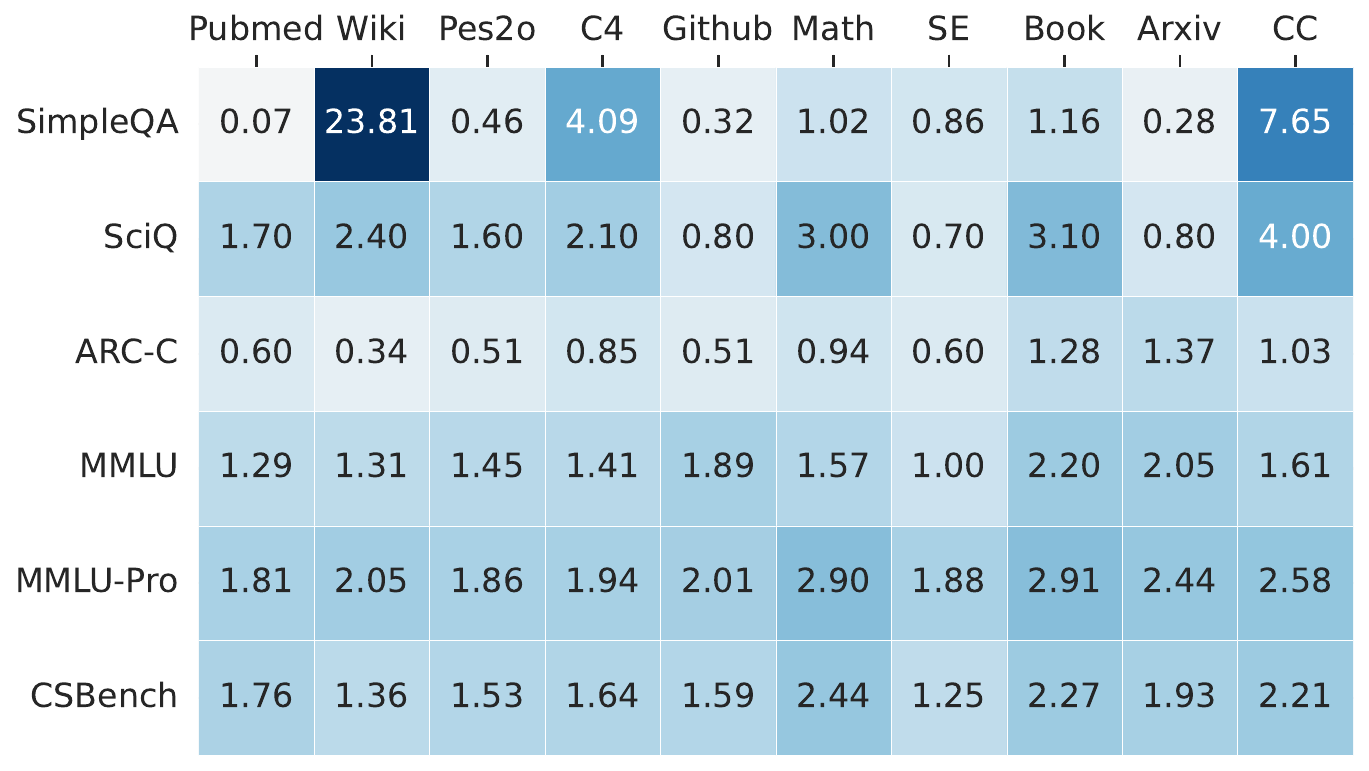}
		\label{fig:LLaMA-3.1-8B}
	}
     \subfigure[Qwen-3-4B]{
		\includegraphics[width=0.31\linewidth]{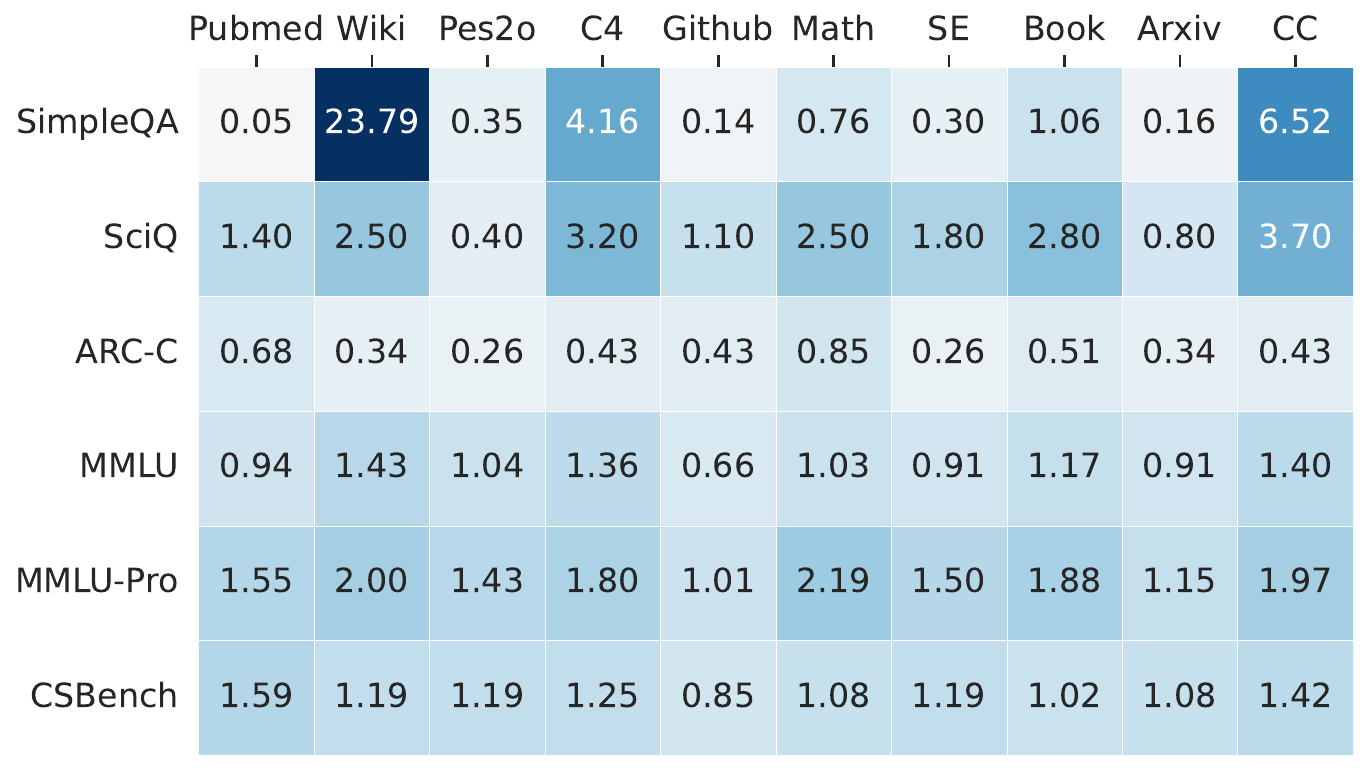}
		\label{fig:qwen3-4b}
	} 
    \subfigure[Qwen-3-8B]{
		\includegraphics[width=0.31\linewidth]{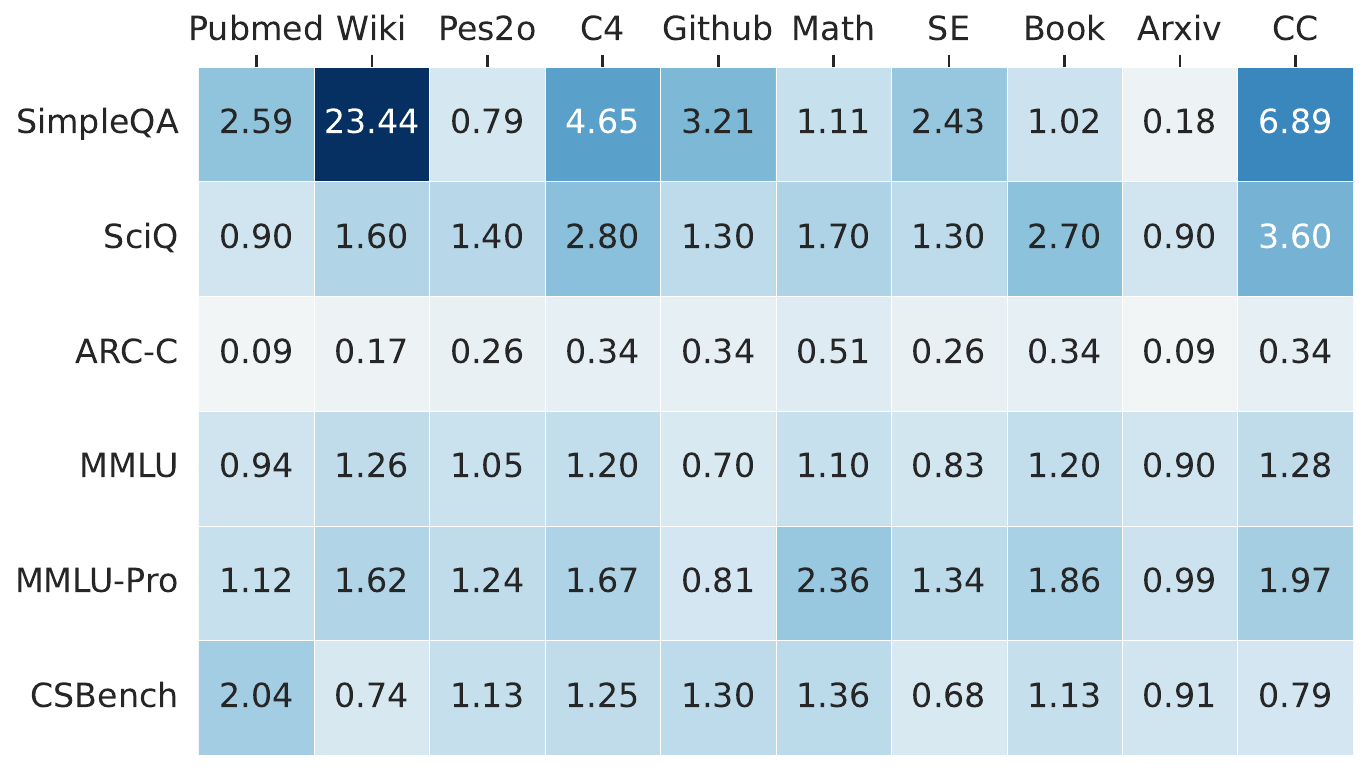}
		\label{fig:qwen3-8b}
	}
    \subfigure[Qwen-3-32B]{
		\includegraphics[width=0.31\linewidth]{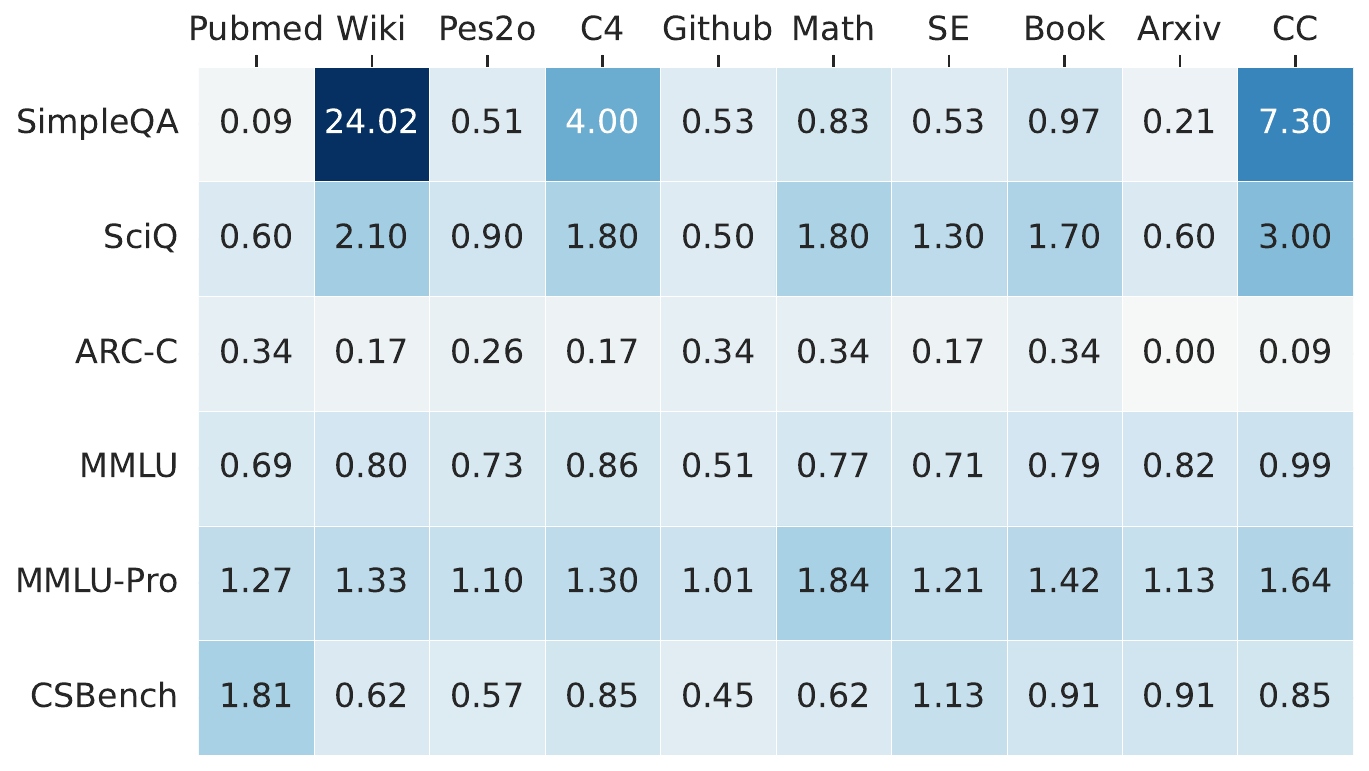}
		\label{fig:Qwen-3-32B}
	} 
        \subfigure[GPT-4o]{
		\includegraphics[width=0.31\linewidth]{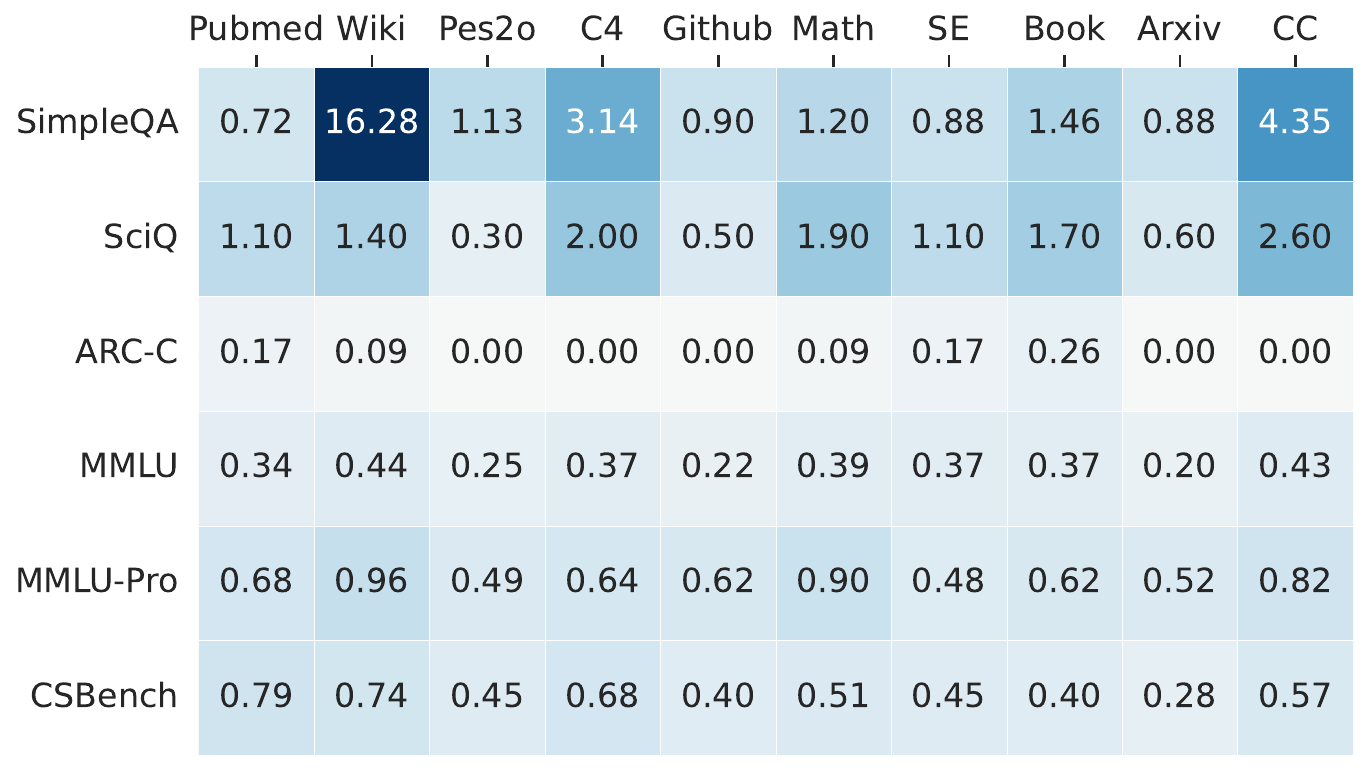}
		\label{fig:gpt-4o}
	}
    \caption{Number of cases (in \%) specifically resolved by retrieving from an individual corpus. \vspace{-1ex}
}
\label{fig:source}
\end{figure*}

\subsection{RQ1: Effectiveness of RAG under Mixture-of-Knowledge Scenarios.}
Figure~\ref{fig:main} illustrates consistent benefits of RAG across a variety of real-world, mixture-of-knowledge scenarios. Notably, \emph{smaller models achieve substantial performance gains}, reflecting their limited capacity to store knowledge internally. In contrast, larger models show diminishing returns from retrieval, with improvements mostly limited to factual knowledge tasks such as SimpleQA. For general and scientific knowledge tasks, which are more effectively captured within the models’ parametric knowledge through pretraining and finetuning, external retrieval brings less benefit.

As some datasets, such as MMLU, cover a wide range of domains, we further analyze the effect of retrieval augmentation across STEM, Social Sciences, Humanities, and Others, as shown in Table~\ref{tab:main}.
We observe consistent trends across all domains: retrieval brings substantial improvements for smaller and mid-sized models, while the gains diminish as model size increases. For the strongest models, retrieval offers marginal or even slightly negative gains, suggesting that larger models are increasingly able to internalize domain knowledge without the need for external retrieval. These results highlight the generalizability of our findings and demonstrate that the impact of retrieval augmentation is robust across different knowledge domains.




\begin{figure*}[t]
	\centering
     \subfigure[SimpleQA]{
		\includegraphics[width=0.315\linewidth]{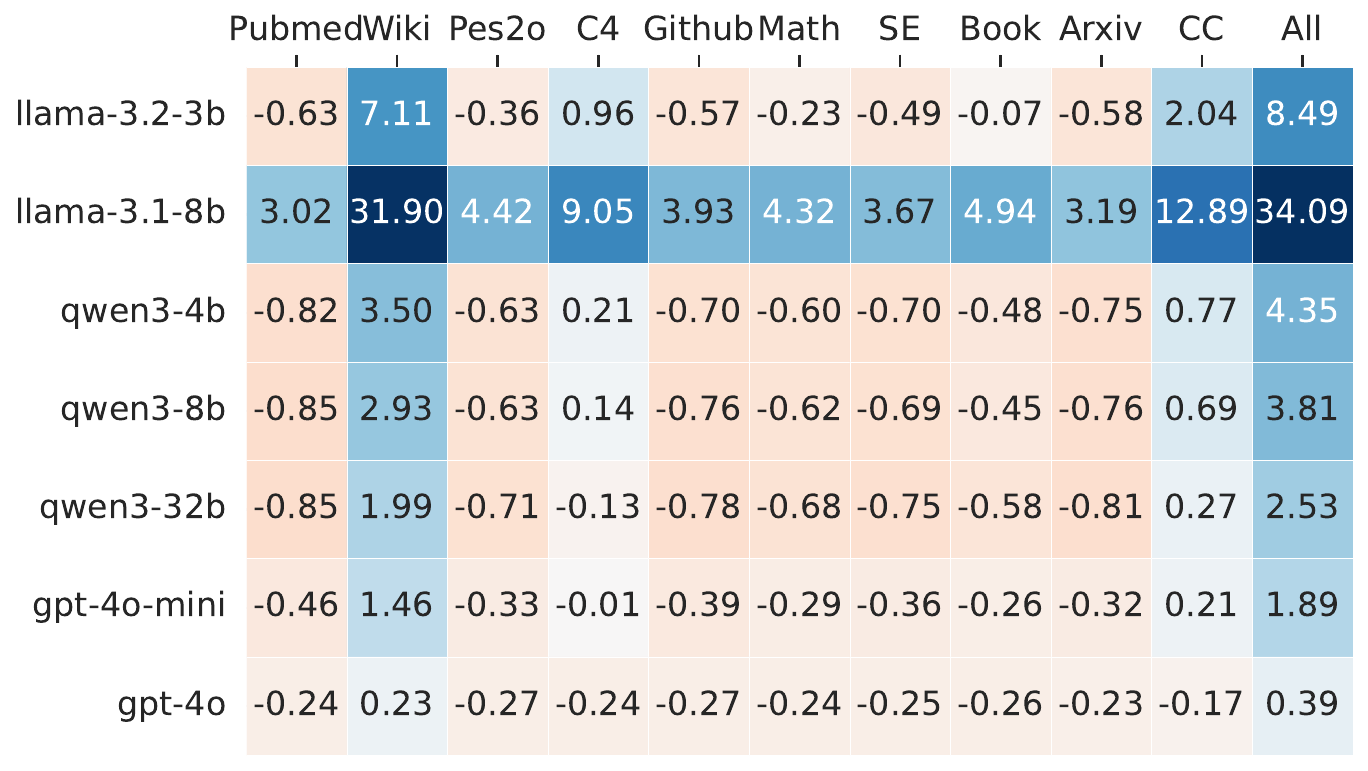}
		\label{fig:SimpleQA-no-wiki-rerank}
	}
     \subfigure[SciQ]{\includegraphics[width=0.315\linewidth]{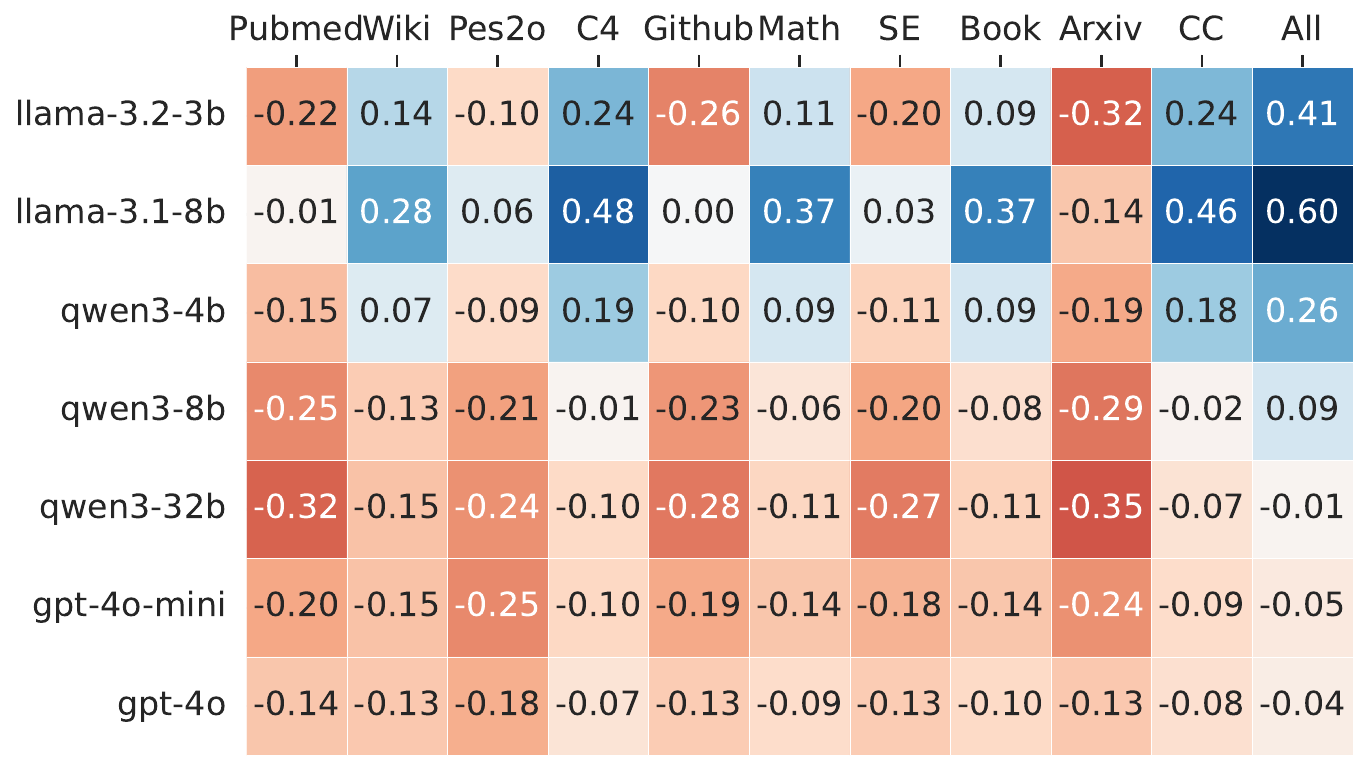}
		\label{fig:Science-rerank}
	} 
    \subfigure[ARC-C]{
		\includegraphics[width=0.315\linewidth]{figures/science-rerank.pdf}
		\label{fig:Science-arcc-rerank}
	} 
     \subfigure[MMLU]{
		\includegraphics[width=0.315\linewidth]{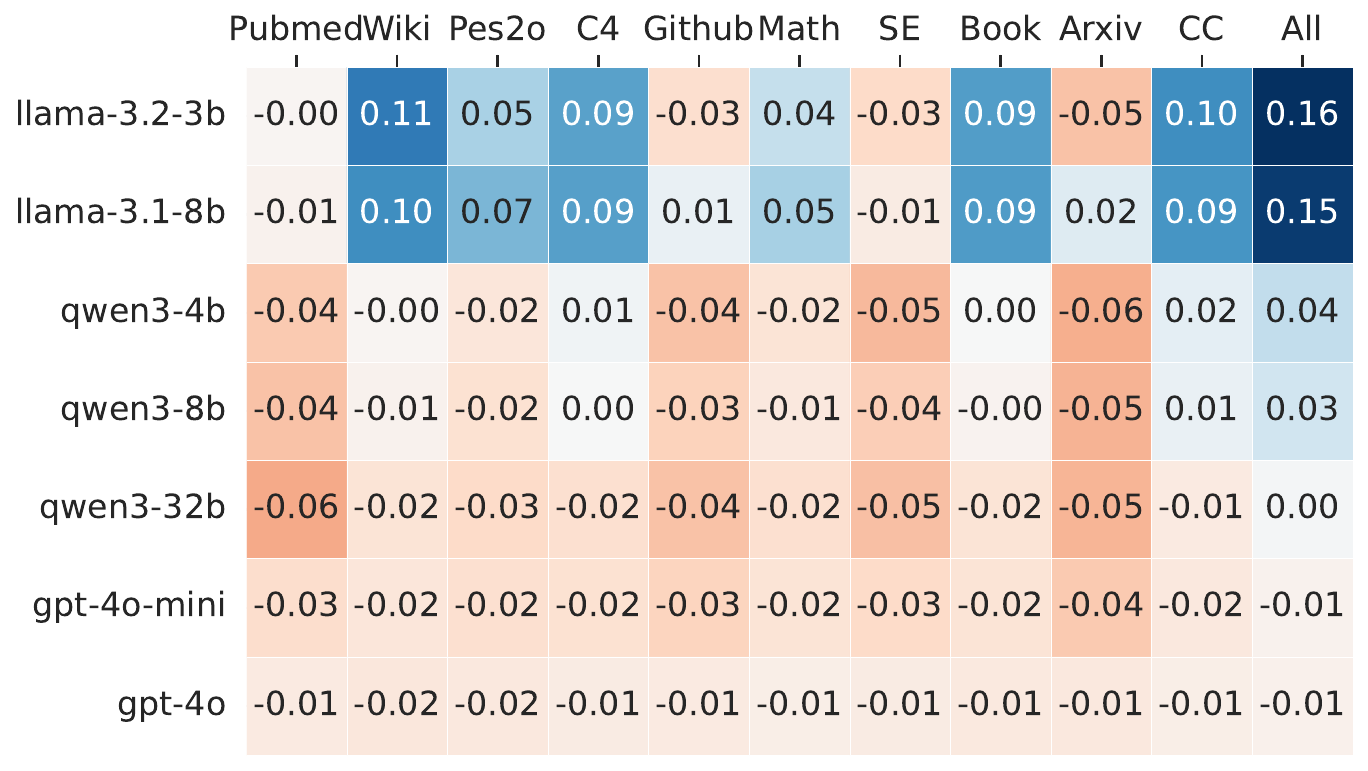}
		\label{fig:mmlu-rerank}
	} 
     \subfigure[MMLU-Pro]{
		\includegraphics[width=0.315\linewidth]{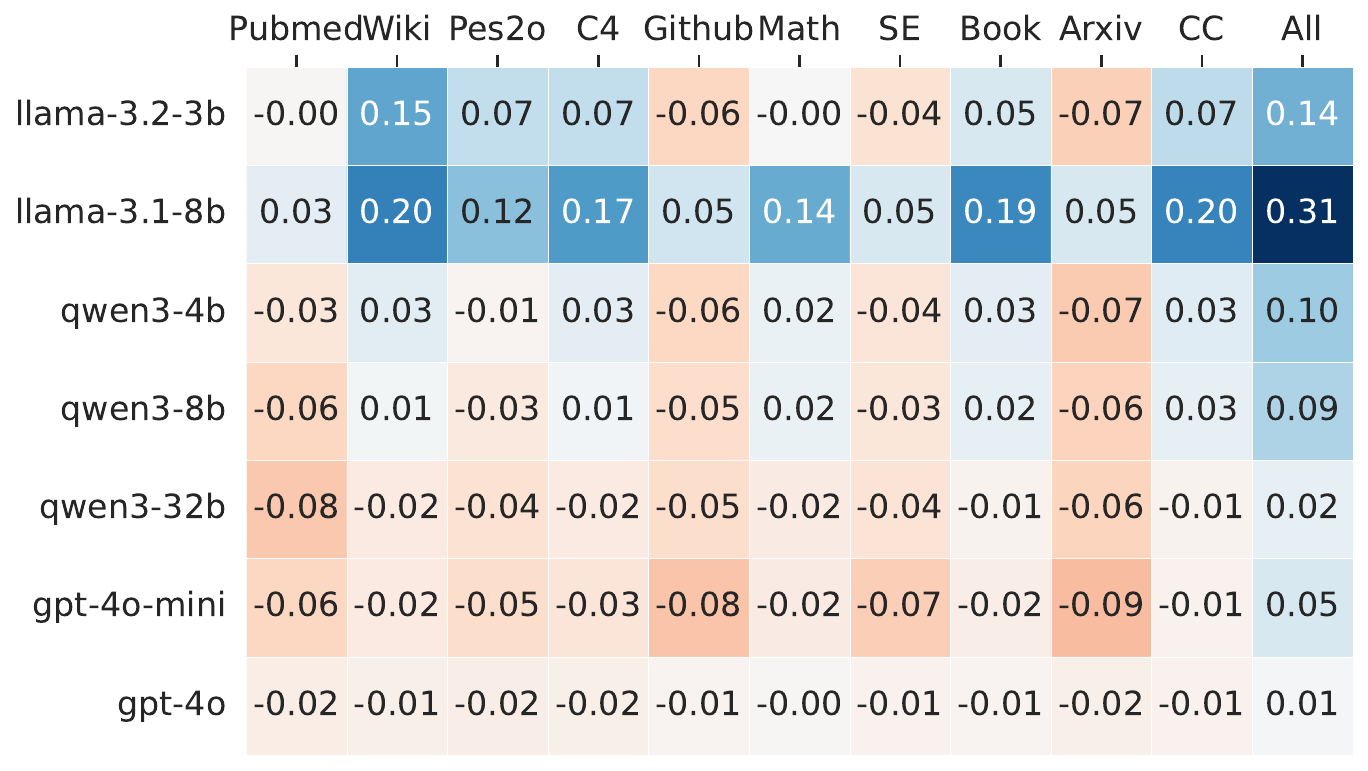}
		\label{fig:mmlu-pro-rerank}
	} 
     \subfigure[CSBench]{
		\includegraphics[width=0.315\linewidth]{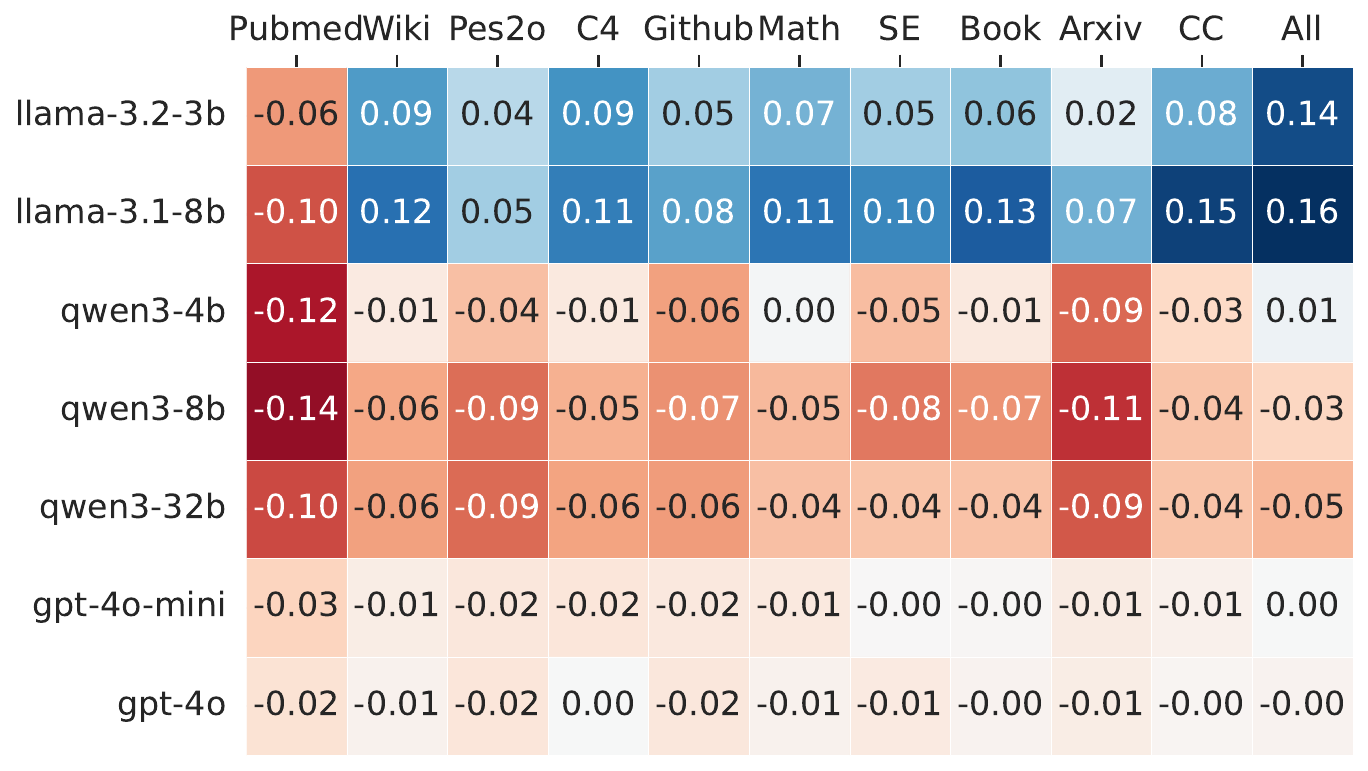}
		\label{fig:csbench-rerank}
	}
    \vspace{-0.5ex}
	\caption{Performance with rerank on retrieval effectiveness across different datasets and models.
}
\label{fig:rerank}
\end{figure*}

\subsection{RQ 2: Instance-level Analysis of Multiple Retrieval Sources}
To investigate whether different retrieval sources provide unique advantages, we conduct an instance-level study (Figure~\ref{fig:source}) measuring the proportion of queries that can be solved only by retrieving from a specific corpus—compared to using all sources or no retrieval at all. 
Our results show that a significant fraction of cases (e.g., 8\%–39\% for Llama-3.1-8B) depend exclusively on retrieval from particular corpora. Crucially, \emph{no single retrieval source consistently outperforms others across all query types}, highlighting the need for dynamic, query-specific routing to the most relevant corpus to maximize RAG performance in heterogeneous, real-world knowledge environments.



\begin{figure*}
    \centering
    \includegraphics[width=\linewidth]{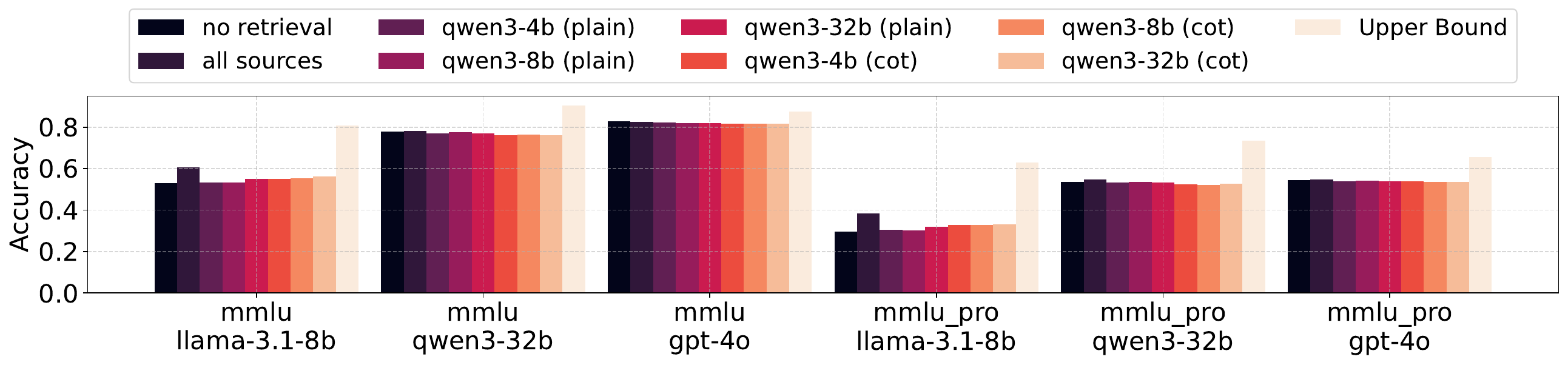}
    \vspace{-1ex}
    \caption{Performance comparison of routing strategies across MMLU and MMLU-Pro datasets. \vspace{-1ex}}
    \label{fig:llmrouting}
\end{figure*}

\subsection{RQ 3: Effectiveness of Reranking}
One possible reason for the limited improvement in RAG performance is the inherent limitations of retrievers. To investigate this, inspired by prior work demonstrating benefits from adding reranking to the RAG pipeline \citep{shao2025scaling,yu2024rankrag}, we apply reranking to the top-30 retrieved results. 
As shown in Figure~\ref{fig:rerank}, reranking yields only marginal gains across datasets. This suggests that \emph{improving retrieval quality  through reranking is insufficient} in mixture-of-knowledge scenarios. The retriever’s limited capacity and restricted access to relevant knowledge highlight the need for deeper integration between knowledge sources, retrieval mechanisms, and generative models.


\subsection{RQ 4: Evaluating LLMs as Routers for Mixture-of-Knowledge Retrieval}

Previous analyses highlight the need for \emph{adaptive retrieval} mechanisms that dynamically route queries to the most relevant corpus based on topic context. Here, we investigate whether current LLMs can effectively perform query routing. 
Figure~\ref{fig:llmrouting} evaluates LLMs from the \texttt{Qwen-3} series (4B, 8B, and 32B parameters) as “routers” that select among heterogeneous knowledge sources at inference time. We compare plain prompting versus chain-of-thought prompting on the MMLU and MMLU-Pro datasets, which cover diverse general and professional domains.

Specifically, we benchmark:
(1) no-retrieval baselines;
(2) static retrieval from all corpora (“all sources”);
(3) LLM-prompted routing variants (plain and chain-of-thought); and
(4) an oracle router upper bound. 
Surprisingly, neither prompting strategy consistently outperforms static retrieval. In fact, both routing approaches often underperform compared to simply retrieving from all sources, and occasionally fall below no-retrieval baselines. Chain-of-thought prompting provides only marginal improvements, while scaling model size yields negligible or even negative returns.

We attribute this failure to two main factors:
(1) \textbf{Inaccurate relevance estimation:} Without dedicated training, LLMs struggle to reliably identify which corpus holds the needed information, especially amid overlapping or stylistically diverse corpora. Minor routing errors propagate to poor retrieval quality.
(2) \textbf{Training-inference mismatch:} LLM training typically lacks explicit multi-source comparison tasks, limiting the effectiveness of prompt-based routing as a meta-reasoning problem. 
Future directions should focus on learned routing modules trained with supervision or reinforcement learning, or on tightly integrated RAG systems that jointly optimize source selection and generation.

\section{Related Work}
Many of prior works on RAG focus on improving either retrievers~\cite{shi-etal-2024-replug,shao-etal-2023-enhancing,xu-etal-2024-bmretriever} or language models~\cite{asai2024selfrag,xu-etal-2024-unsupervised,huang2024raven}, as well as optimizing the whole RAG pipeline~\cite{trivedi-etal-2023-interleaving,wang-etal-2024-blendfilter}.  To better understand the impact of retrieval corpora, \citet{shao2025scaling} explores how corpus scale influences RAG performance, while \citet{cuconasu2024power,niu-etal-2024-ragtruth} examine the effects of noisy or imperfect corpora. \citet{chen2025towards,xu-etal-2024-ram,medrag} assess RAG systems across different knowledge sources, though they primarily focus on biomedical tasks. In contrast, our work provides a broader investigation, spanning diverse domains, retrieval corpora, and LLM backbones.

In parallel, data routing strategies have been 
explored for LLMs. 
\citet{ong2025routellm,frick2025prompt} routes each prompt to a specialized LLM. Within the RAG setting, \citet{mu-etal-2024-query,mu2025unsupervised} investigate routing prompts to different search engines, \citet{zhang2025query} routes queries to different LLMs, 
and \citet{wu2025self,asai2024selfrag,yao2024seakr} studied \emph{adaptive retrieval} to dynamically determine retrieval during inference. 
Distinct from these efforts, we also focuses on routing but operates at the \emph{corpus level}, studying adaptive selection among heterogeneous knowledge sources.

\section{Conclusion}
Our comprehensive empirical analysis highlights key challenges in deploying RAG systems within realistic, heterogeneous knowledge environments. While retrieval offers benefits for smaller LLMs, larger models exhibit only marginal improvements, except for factual tasks. Additionally, reranking techniques and prompt-based knowledge routing provide limited gains, justifying the need for future research focused on adaptive routing and tighter integration between retrieval and generation.

We identify several promising directions to address these challenges:
(1) \emph{Reasoning-enhanced search}: \citet{shao2025reasonir,zhuang2025rank} incorporate reasoning directly into retrieval and ranking stages, improving relevance at the expense of increased inference time.
(2) \emph{Agentic RAG systems}: Query rewriting \citep{ma2023query} and decomposition \citep{li2025search,jin2025search} enable multi-turn retrieval and generation workflows to better handle complex queries.

\section*{Limitations}
Our study primarily targets question answering tasks with short-form answers, which may limit the applicability of our results to other settings such as open-ended generation or long-form reasoning. While we evaluate several widely used models, our experiments do not comprehensively cover larger open-source models (e.g., DeepSeek-V3~\citep{liu2024deepseek} or Llama-4~\citep{meta2025llama}) or alternative retrieval paradigms, mainly due to computational constraints. Consequently, there remains ample opportunity for future work to explore these directions. Lastly, we do not assess computational efficiency or latency trade-offs, which are critical considerations for real-world deployment of RAG systems but are beyond the scope of this study.




\bibliography{custom}

\begin{thebibliography}{51}
\providecommand{\natexlab}[1]{#1}

\bibitem[{Asai et~al.(2024)Asai, Wu, Wang, Sil, and Hajishirzi}]{asai2024selfrag}
Akari Asai, Zeqiu Wu, Yizhong Wang, Avirup Sil, and Hannaneh Hajishirzi. 2024.
\newblock \href {https://openreview.net/forum?id=hSyW5go0v8} {Self-{RAG}: Learning to retrieve, generate, and critique through self-reflection}.
\newblock In \emph{The Twelfth International Conference on Learning Representations}.

\bibitem[{Chen et~al.(2024)Chen, Xiao, Zhang, Luo, Lian, and Liu}]{chen2024bge}
Jianlv Chen, Shitao Xiao, Peitian Zhang, Kun Luo, Defu Lian, and Zheng Liu. 2024.
\newblock Bge m3-embedding: Multi-lingual, multi-functionality, multi-granularity text embeddings through self-knowledge distillation.
\newblock \emph{arXiv preprint arXiv:2402.03216}.

\bibitem[{Chen et~al.(2025)Chen, Liao, Jiang, Wang, Guo, Wang, and Wang}]{chen2025towards}
Zhe Chen, Yusheng Liao, Shuyang Jiang, Pingjie Wang, Yiqiu Guo, Yanfeng Wang, and Yu~Wang. 2025.
\newblock Towards omni-rag: Comprehensive retrieval-augmented generation for large language models in medical applications.
\newblock \emph{arXiv preprint arXiv:2501.02460}.

\bibitem[{Clark et~al.(2018)Clark, Cowhey, Etzioni, Khot, Sabharwal, Schoenick, and Tafjord}]{arcc}
Peter Clark, Isaac Cowhey, Oren Etzioni, Tushar Khot, Ashish Sabharwal, Carissa Schoenick, and Oyvind Tafjord. 2018.
\newblock Think you have solved question answering? try arc, the ai2 reasoning challenge.
\newblock \emph{arXiv preprint arXiv:1803.05457}.

\bibitem[{Cuconasu et~al.(2024)Cuconasu, Trappolini, Siciliano, Filice, Campagnano, Maarek, Tonellotto, and Silvestri}]{cuconasu2024power}
Florin Cuconasu, Giovanni Trappolini, Federico Siciliano, Simone Filice, Cesare Campagnano, Yoelle Maarek, Nicola Tonellotto, and Fabrizio Silvestri. 2024.
\newblock The power of noise: Redefining retrieval for rag systems.
\newblock In \emph{Proceedings of the 47th International ACM SIGIR Conference on Research and Development in Information Retrieval}, pages 719--729.

\bibitem[{Frick et~al.(2025)Frick, Chen, Tennyson, Li, Chiang, Angelopoulos, and Stoica}]{frick2025prompt}
Evan Frick, Connor Chen, Joseph Tennyson, Tianle Li, Wei-Lin Chiang, Anastasios~N Angelopoulos, and Ion Stoica. 2025.
\newblock Prompt-to-leaderboard.
\newblock \emph{arXiv preprint arXiv:2502.14855}.

\bibitem[{Grattafiori et~al.(2024)Grattafiori, Dubey, Jauhri, Pandey, Kadian, Al-Dahle, Letman, Mathur, Schelten, Vaughan et~al.}]{grattafiori2024llama}
Aaron Grattafiori, Abhimanyu Dubey, Abhinav Jauhri, Abhinav Pandey, Abhishek Kadian, Ahmad Al-Dahle, Aiesha Letman, Akhil Mathur, Alan Schelten, Alex Vaughan, et~al. 2024.
\newblock The llama 3 herd of models.
\newblock \emph{arXiv preprint arXiv:2407.21783}.

\bibitem[{Hendrycks et~al.(2021)Hendrycks, Burns, Basart, Zou, Mazeika, Song, and Steinhardt}]{mmlu}
Dan Hendrycks, Collin Burns, Steven Basart, Andy Zou, Mantas Mazeika, Dawn Song, and Jacob Steinhardt. 2021.
\newblock \href {https://openreview.net/forum?id=d7KBjmI3GmQ} {Measuring massive multitask language understanding}.
\newblock In \emph{International Conference on Learning Representations}.

\bibitem[{Huang et~al.(2024)Huang, Ping, Xu, Shoeybi, Chang, and Catanzaro}]{huang2024raven}
Jie Huang, Wei Ping, Peng Xu, Mohammad Shoeybi, Kevin Chang, and Bryan Catanzaro. 2024.
\newblock \href {https://openreview.net/forum?id=GMalvQu0XL} {{RAVEN}: In-context learning with retrieval-augmented encoder-decoder language models}.
\newblock In \emph{First Conference on Language Modeling}.

\bibitem[{Hurst et~al.(2024)Hurst, Lerer, Goucher, Perelman, Ramesh, Clark, Ostrow, Welihinda, Hayes, Radford et~al.}]{hurst2024gpt4o}
Aaron Hurst, Adam Lerer, Adam~P Goucher, Adam Perelman, Aditya Ramesh, Aidan Clark, AJ~Ostrow, Akila Welihinda, Alan Hayes, Alec Radford, et~al. 2024.
\newblock Gpt-4o system card.
\newblock \emph{arXiv preprint arXiv:2410.21276}.

\bibitem[{Islam et~al.(2024)Islam, Rahman, Hossain, Hoque, Joty, and Parvez}]{islam-etal-2024-open}
Shayekh~Bin Islam, Md~Asib Rahman, K~S M~Tozammel Hossain, Enamul Hoque, Shafiq Joty, and Md~Rizwan Parvez. 2024.
\newblock \href {https://doi.org/10.18653/v1/2024.findings-emnlp.831} {Open-{RAG}: Enhanced retrieval augmented reasoning with open-source large language models}.
\newblock In \emph{Findings of the Association for Computational Linguistics: EMNLP 2024}, pages 14231--14244, Miami, Florida, USA. Association for Computational Linguistics.

\bibitem[{Izacard et~al.(2023)Izacard, Lewis, Lomeli, Hosseini, Petroni, Schick, Dwivedi-Yu, Joulin, Riedel, and Grave}]{izacard2023atlas}
Gautier Izacard, Patrick Lewis, Maria Lomeli, Lucas Hosseini, Fabio Petroni, Timo Schick, Jane Dwivedi-Yu, Armand Joulin, Sebastian Riedel, and Edouard Grave. 2023.
\newblock Atlas: Few-shot learning with retrieval augmented language models.
\newblock \emph{Journal of Machine Learning Research}, 24(251):1--43.

\bibitem[{Jin et~al.(2025)Jin, Zeng, Yue, Yoon, Arik, Wang, Zamani, and Han}]{jin2025search}
Bowen Jin, Hansi Zeng, Zhenrui Yue, Jinsung Yoon, Sercan Arik, Dong Wang, Hamed Zamani, and Jiawei Han. 2025.
\newblock Search-r1: Training llms to reason and leverage search engines with reinforcement learning.
\newblock \emph{arXiv preprint arXiv:2503.09516}.

\bibitem[{Joshi et~al.(2017)Joshi, Choi, Weld, and Zettlemoyer}]{joshi-etal-2017-triviaqa}
Mandar Joshi, Eunsol Choi, Daniel Weld, and Luke Zettlemoyer. 2017.
\newblock \href {https://doi.org/10.18653/v1/P17-1147} {{T}rivia{QA}: A large scale distantly supervised challenge dataset for reading comprehension}.
\newblock In \emph{Proceedings of the 55th Annual Meeting of the Association for Computational Linguistics (Volume 1: Long Papers)}, pages 1601--1611, Vancouver, Canada. Association for Computational Linguistics.

\bibitem[{Kwiatkowski et~al.(2019)Kwiatkowski, Palomaki, Redfield, Collins, Parikh, Alberti, Epstein, Polosukhin, Devlin, Lee, Toutanova, Jones, Kelcey, Chang, Dai, Uszkoreit, Le, and Petrov}]{nq}
Tom Kwiatkowski, Jennimaria Palomaki, Olivia Redfield, Michael Collins, Ankur Parikh, Chris Alberti, Danielle Epstein, Illia Polosukhin, Jacob Devlin, Kenton Lee, Kristina Toutanova, Llion Jones, Matthew Kelcey, Ming-Wei Chang, Andrew~M. Dai, Jakob Uszkoreit, Quoc Le, and Slav Petrov. 2019.
\newblock \href {https://aclanthology.org/Q19-1026} {Natural questions: A benchmark for question answering research}.
\newblock \emph{Transactions of the Association for Computational Linguistics}, 7:452--466.

\bibitem[{Lewis et~al.(2020)Lewis, Perez, Piktus, Petroni, Karpukhin, Goyal, K{\"u}ttler, Lewis, Yih, Rockt{\"a}schel et~al.}]{lewis2020retrieval}
Patrick Lewis, Ethan Perez, Aleksandra Piktus, Fabio Petroni, Vladimir Karpukhin, Naman Goyal, Heinrich K{\"u}ttler, Mike Lewis, Wen-tau Yih, Tim Rockt{\"a}schel, et~al. 2020.
\newblock Retrieval-augmented generation for knowledge-intensive nlp tasks.
\newblock \emph{Advances in neural information processing systems}, 33:9459--9474.

\bibitem[{Li et~al.(2025)Li, Dong, Jin, Zhang, Zhou, Zhu, Zhang, and Dou}]{li2025search}
Xiaoxi Li, Guanting Dong, Jiajie Jin, Yuyao Zhang, Yujia Zhou, Yutao Zhu, Peitian Zhang, and Zhicheng Dou. 2025.
\newblock Search-o1: Agentic search-enhanced large reasoning models.
\newblock \emph{arXiv preprint arXiv:2501.05366}.

\bibitem[{Lin et~al.(2024)Lin, Chen, Chen, Shi, Lomeli, James, Rodriguez, Kahn, Szilvasy, Lewis, Zettlemoyer, and tau Yih}]{lin2024radit}
Xi~Victoria Lin, Xilun Chen, Mingda Chen, Weijia Shi, Maria Lomeli, Richard James, Pedro Rodriguez, Jacob Kahn, Gergely Szilvasy, Mike Lewis, Luke Zettlemoyer, and Wen tau Yih. 2024.
\newblock \href {https://openreview.net/forum?id=22OTbutug9} {{RA}-{DIT}: Retrieval-augmented dual instruction tuning}.
\newblock In \emph{The Twelfth International Conference on Learning Representations}.

\bibitem[{Liu et~al.(2024)Liu, Feng, Xue, Wang, Wu, Lu, Zhao, Deng, Zhang, Ruan et~al.}]{liu2024deepseek}
Aixin Liu, Bei Feng, Bing Xue, Bingxuan Wang, Bochao Wu, Chengda Lu, Chenggang Zhao, Chengqi Deng, Chenyu Zhang, Chong Ruan, et~al. 2024.
\newblock Deepseek-v3 technical report.
\newblock \emph{arXiv preprint arXiv:2412.19437}.

\bibitem[{Ma et~al.(2023)Ma, Gong, He, hai zhao, and Duan}]{ma2023query}
Xinbei Ma, Yeyun Gong, Pengcheng He, hai zhao, and Nan Duan. 2023.
\newblock \href {https://openreview.net/forum?id=gXq1cwkUZc} {Query rewriting in retrieval-augmented large language models}.
\newblock In \emph{The 2023 Conference on Empirical Methods in Natural Language Processing}.

\bibitem[{Mallen et~al.(2023)Mallen, Asai, Zhong, Das, Khashabi, and Hajishirzi}]{mallen-etal-2023-trust}
Alex Mallen, Akari Asai, Victor Zhong, Rajarshi Das, Daniel Khashabi, and Hannaneh Hajishirzi. 2023.
\newblock \href {https://doi.org/10.18653/v1/2023.acl-long.546} {When not to trust language models: Investigating effectiveness of parametric and non-parametric memories}.
\newblock In \emph{Proceedings of the 61st Annual Meeting of the Association for Computational Linguistics (Volume 1: Long Papers)}, pages 9802--9822, Toronto, Canada. Association for Computational Linguistics.

\bibitem[{Meta(2025)}]{meta2025llama}
AI~Meta. 2025.
\newblock \href {https://ai.meta.com/blog/llama-4-multimodal-intelligence/} {The llama 4 herd: The beginning of a new era of natively multimodal ai innovation}.

\bibitem[{Mu et~al.(2024)Mu, Jiang, Zhang, Liuchu, Li, Xie, and Huang}]{mu-etal-2024-query}
Feiteng Mu, Yong Jiang, Liwen Zhang, Liuchu Liuchu, Wenjie Li, Pengjun Xie, and Fei Huang. 2024.
\newblock \href {https://doi.org/10.18653/v1/2024.findings-emnlp.598} {Query routing for homogeneous tools: An instantiation in the {RAG} scenario}.
\newblock In \emph{Findings of the Association for Computational Linguistics: EMNLP 2024}, pages 10225--10230, Miami, Florida, USA. Association for Computational Linguistics.

\bibitem[{Mu et~al.(2025)Mu, Zhang, Jiang, Li, Zhang, Xie, and Huang}]{mu2025unsupervised}
Feiteng Mu, Liwen Zhang, Yong Jiang, Wenjie Li, Zhen Zhang, Pengjun Xie, and Fei Huang. 2025.
\newblock Unsupervised query routing for retrieval augmented generation.
\newblock \emph{arXiv preprint arXiv:2501.07793}.

\bibitem[{Niu et~al.(2024)Niu, Wu, Zhu, Xu, Shum, Zhong, Song, and Zhang}]{niu-etal-2024-ragtruth}
Cheng Niu, Yuanhao Wu, Juno Zhu, Siliang Xu, KaShun Shum, Randy Zhong, Juntong Song, and Tong Zhang. 2024.
\newblock \href {https://doi.org/10.18653/v1/2024.acl-long.585} {{RAGT}ruth: A hallucination corpus for developing trustworthy retrieval-augmented language models}.
\newblock In \emph{Proceedings of the 62nd Annual Meeting of the Association for Computational Linguistics (Volume 1: Long Papers)}, pages 10862--10878, Bangkok, Thailand. Association for Computational Linguistics.

\bibitem[{Ong et~al.(2025)Ong, Almahairi, Wu, Chiang, Wu, Gonzalez, Kadous, and Stoica}]{ong2025routellm}
Isaac Ong, Amjad Almahairi, Vincent Wu, Wei-Lin Chiang, Tianhao Wu, Joseph~E. Gonzalez, M~Waleed Kadous, and Ion Stoica. 2025.
\newblock \href {https://openreview.net/forum?id=8sSqNntaMr} {Route{LLM}: Learning to route {LLM}s from preference data}.
\newblock In \emph{The Thirteenth International Conference on Learning Representations}.

\bibitem[{Petroni et~al.(2021)Petroni, Piktus, Fan, Lewis, Yazdani, De~Cao, Thorne, Jernite, Karpukhin, Maillard, Plachouras, Rockt{\"a}schel, and Riedel}]{petroni-etal-2021-kilt}
Fabio Petroni, Aleksandra Piktus, Angela Fan, Patrick Lewis, Majid Yazdani, Nicola De~Cao, James Thorne, Yacine Jernite, Vladimir Karpukhin, Jean Maillard, Vassilis Plachouras, Tim Rockt{\"a}schel, and Sebastian Riedel. 2021.
\newblock \href {https://doi.org/10.18653/v1/2021.naacl-main.200} {{KILT}: a benchmark for knowledge intensive language tasks}.
\newblock In \emph{Proceedings of the 2021 Conference of the North American Chapter of the Association for Computational Linguistics: Human Language Technologies}, pages 2523--2544, Online. Association for Computational Linguistics.

\bibitem[{Shao et~al.(2024)Shao, He, Asai, Shi, Dettmers, Min, Zettlemoyer, and Koh}]{shao2025scaling}
Rulin Shao, Jacqueline He, Akari Asai, Weijia Shi, Tim Dettmers, Sewon Min, Luke Zettlemoyer, and Pang Wei~W Koh. 2024.
\newblock Scaling retrieval-based language models with a trillion-token datastore.
\newblock \emph{Advances in Neural Information Processing Systems}, 37:91260--91299.

\bibitem[{Shao et~al.(2025)Shao, Qiao, Kishore, Muennighoff, Lin, Rus, Low, Min, Yih, Koh et~al.}]{shao2025reasonir}
Rulin Shao, Rui Qiao, Varsha Kishore, Niklas Muennighoff, Xi~Victoria Lin, Daniela Rus, Bryan Kian~Hsiang Low, Sewon Min, Wen-tau Yih, Pang~Wei Koh, et~al. 2025.
\newblock Reasonir: Training retrievers for reasoning tasks.
\newblock \emph{arXiv preprint arXiv:2504.20595}.

\bibitem[{Shao et~al.(2023)Shao, Gong, Shen, Huang, Duan, and Chen}]{shao-etal-2023-enhancing}
Zhihong Shao, Yeyun Gong, Yelong Shen, Minlie Huang, Nan Duan, and Weizhu Chen. 2023.
\newblock \href {https://doi.org/10.18653/v1/2023.findings-emnlp.620} {Enhancing retrieval-augmented large language models with iterative retrieval-generation synergy}.
\newblock In \emph{Findings of the Association for Computational Linguistics: EMNLP 2023}, pages 9248--9274, Singapore. Association for Computational Linguistics.

\bibitem[{Shi et~al.(2024)Shi, Min, Yasunaga, Seo, James, Lewis, Zettlemoyer, and Yih}]{shi-etal-2024-replug}
Weijia Shi, Sewon Min, Michihiro Yasunaga, Minjoon Seo, Richard James, Mike Lewis, Luke Zettlemoyer, and Wen-tau Yih. 2024.
\newblock \href {https://doi.org/10.18653/v1/2024.naacl-long.463} {{REPLUG}: Retrieval-augmented black-box language models}.
\newblock In \emph{Proceedings of the 2024 Conference of the North American Chapter of the Association for Computational Linguistics: Human Language Technologies (Volume 1: Long Papers)}, pages 8371--8384, Mexico City, Mexico. Association for Computational Linguistics.

\bibitem[{Song et~al.(2025)Song, Diao, Dong, Wang, Fu, Qiao, Wang, Fu, Wu, Liang, Zeng, Wang, GongQue, Yu, Tan, and Xu}]{song2025csbench}
Xiaoshuai Song, Muxi Diao, Guanting Dong, Zhengyang Wang, Yujia Fu, Runqi Qiao, Zhexu Wang, Dayuan Fu, Huangxuan Wu, Bin Liang, Weihao Zeng, Yejie Wang, Zhuoma GongQue, Jianing Yu, Qiuna Tan, and Weiran Xu. 2025.
\newblock \href {https://openreview.net/forum?id=fjEZ2LPceZ} {{CS}-bench: A comprehensive benchmark for large language models towards computer science mastery}.
\newblock In \emph{The Thirteenth International Conference on Learning Representations}.

\bibitem[{Thorne et~al.(2018)Thorne, Vlachos, Christodoulopoulos, and Mittal}]{fever}
James Thorne, Andreas Vlachos, Christos Christodoulopoulos, and Arpit Mittal. 2018.
\newblock \href {https://aclanthology.org/N18-1074} {{FEVER}: a large-scale dataset for fact extraction and {VER}ification}.
\newblock In \emph{Proceedings of the 2018 Conference of the North {A}merican Chapter of the Association for Computational Linguistics: Human Language Technologies, Volume 1 (Long Papers)}, pages 809--819, New Orleans, Louisiana. Association for Computational Linguistics.

\bibitem[{Trivedi et~al.(2023)Trivedi, Balasubramanian, Khot, and Sabharwal}]{trivedi-etal-2023-interleaving}
Harsh Trivedi, Niranjan Balasubramanian, Tushar Khot, and Ashish Sabharwal. 2023.
\newblock \href {https://doi.org/10.18653/v1/2023.acl-long.557} {Interleaving retrieval with chain-of-thought reasoning for knowledge-intensive multi-step questions}.
\newblock In \emph{Proceedings of the 61st Annual Meeting of the Association for Computational Linguistics (Volume 1: Long Papers)}, pages 10014--10037, Toronto, Canada. Association for Computational Linguistics.

\bibitem[{Wang et~al.(2024{\natexlab{a}})Wang, Li, Jiang, Tian, Wang, Luo, Tang, Cheng, Zhao, and Gao}]{wang-etal-2024-blendfilter}
Haoyu Wang, Ruirui Li, Haoming Jiang, Jinjin Tian, Zhengyang Wang, Chen Luo, Xianfeng Tang, Monica~Xiao Cheng, Tuo Zhao, and Jing Gao. 2024{\natexlab{a}}.
\newblock \href {https://doi.org/10.18653/v1/2024.emnlp-main.58} {{B}lend{F}ilter: Advancing retrieval-augmented large language models via query generation blending and knowledge filtering}.
\newblock In \emph{Proceedings of the 2024 Conference on Empirical Methods in Natural Language Processing}, pages 1009--1025, Miami, Florida, USA. Association for Computational Linguistics.

\bibitem[{Wang et~al.(2024{\natexlab{b}})Wang, Ma, Zhang, Ni, Chandra, Guo, Ren, Arulraj, He, Jiang, Li, Ku, Wang, Zhuang, Fan, Yue, and Chen}]{mmlupro}
Yubo Wang, Xueguang Ma, Ge~Zhang, Yuansheng Ni, Abhranil Chandra, Shiguang Guo, Weiming Ren, Aaran Arulraj, Xuan He, Ziyan Jiang, Tianle Li, Max Ku, Kai Wang, Alex Zhuang, Rongqi Fan, Xiang Yue, and Wenhu Chen. 2024{\natexlab{b}}.
\newblock \href {https://openreview.net/forum?id=y10DM6R2r3} {{MMLU}-pro: A more robust and challenging multi-task language understanding benchmark}.
\newblock In \emph{The Thirty-eight Conference on Neural Information Processing Systems Datasets and Benchmarks Track}.

\bibitem[{Wei et~al.(2024)Wei, Karina, Chung, Jiao, Papay, Glaese, Schulman, and Fedus}]{wei2024measuring}
Jason Wei, Nguyen Karina, Hyung~Won Chung, Yunxin~Joy Jiao, Spencer Papay, Amelia Glaese, John Schulman, and William Fedus. 2024.
\newblock Measuring short-form factuality in large language models.
\newblock \emph{arXiv preprint arXiv:2411.04368}.

\bibitem[{Wei et~al.(2022)Wei, Wang, Schuurmans, Bosma, Xia, Chi, Le, Zhou et~al.}]{wei2022chain}
Jason Wei, Xuezhi Wang, Dale Schuurmans, Maarten Bosma, Fei Xia, Ed~Chi, Quoc~V Le, Denny Zhou, et~al. 2022.
\newblock Chain-of-thought prompting elicits reasoning in large language models.
\newblock \emph{Advances in neural information processing systems}, 35:24824--24837.

\bibitem[{Welbl et~al.(2017)Welbl, Liu, and Gardner}]{welbl2017crowdsourcing}
Johannes Welbl, Nelson~F Liu, and Matt Gardner. 2017.
\newblock Crowdsourcing multiple choice science questions.
\newblock \emph{arXiv preprint arXiv:1707.06209}.

\bibitem[{Wu et~al.(2025)Wu, Gu, Chang, and Peng}]{wu2025self}
Di~Wu, Jia-Chen Gu, Kai-Wei Chang, and Nanyun Peng. 2025.
\newblock Self-routing rag: Binding selective retrieval with knowledge verbalization.
\newblock \emph{arXiv preprint arXiv:2504.01018}.

\bibitem[{Xiong et~al.(2024)Xiong, Jin, Lu, and Zhang}]{medrag}
Guangzhi Xiong, Qiao Jin, Zhiyong Lu, and Aidong Zhang. 2024.
\newblock \href {https://doi.org/10.18653/v1/2024.findings-acl.372} {Benchmarking retrieval-augmented generation for medicine}.
\newblock In \emph{Findings of the Association for Computational Linguistics: ACL 2024}, pages 6233--6251, Bangkok, Thailand. Association for Computational Linguistics.

\bibitem[{Xu et~al.(2024{\natexlab{a}})Xu, Shi, Yu, Zhuang, Jin, Wang, Ho, and Yang}]{xu-etal-2024-ram}
Ran Xu, Wenqi Shi, Yue Yu, Yuchen Zhuang, Bowen Jin, May~Dongmei Wang, Joyce Ho, and Carl Yang. 2024{\natexlab{a}}.
\newblock \href {https://doi.org/10.18653/v1/2024.acl-short.68} {{RAM}-{EHR}: Retrieval augmentation meets clinical predictions on electronic health records}.
\newblock In \emph{Proceedings of the 62nd Annual Meeting of the Association for Computational Linguistics (Volume 2: Short Papers)}, pages 754--765, Bangkok, Thailand. Association for Computational Linguistics.

\bibitem[{Xu et~al.(2024{\natexlab{b}})Xu, Shi, Yu, Zhuang, Zhu, Wang, Ho, Zhang, and Yang}]{xu-etal-2024-bmretriever}
Ran Xu, Wenqi Shi, Yue Yu, Yuchen Zhuang, Yanqiao Zhu, May~Dongmei Wang, Joyce~C. Ho, Chao Zhang, and Carl Yang. 2024{\natexlab{b}}.
\newblock \href {https://doi.org/10.18653/v1/2024.emnlp-main.1241} {Bmretriever: Tuning large language models as better biomedical text retrievers}.
\newblock In \emph{Proceedings of the 2024 Conference on Empirical Methods in Natural Language Processing}, pages 22234--22254, Miami, Florida, USA. Association for Computational Linguistics.

\bibitem[{Xu et~al.(2024{\natexlab{c}})Xu, Pang, Yu, Meng, Shen, Cheng, and Zhou}]{xu-etal-2024-unsupervised}
Shicheng Xu, Liang Pang, Mo~Yu, Fandong Meng, Huawei Shen, Xueqi Cheng, and Jie Zhou. 2024{\natexlab{c}}.
\newblock \href {https://doi.org/10.18653/v1/2024.acl-long.9} {Unsupervised information refinement training of large language models for retrieval-augmented generation}.
\newblock In \emph{Proceedings of the 62nd Annual Meeting of the Association for Computational Linguistics (Volume 1: Long Papers)}, pages 133--145, Bangkok, Thailand. Association for Computational Linguistics.

\bibitem[{Yang et~al.(2025)Yang, Li, Yang, Zhang, Hui, Zheng, Yu, Gao, Huang, Lv et~al.}]{yang2025qwen3}
An~Yang, Anfeng Li, Baosong Yang, Beichen Zhang, Binyuan Hui, Bo~Zheng, Bowen Yu, Chang Gao, Chengen Huang, Chenxu Lv, et~al. 2025.
\newblock Qwen3 technical report.
\newblock \emph{arXiv preprint arXiv:2505.09388}.

\bibitem[{Yang et~al.(2018)Yang, Qi, Zhang, Bengio, Cohen, Salakhutdinov, and Manning}]{hotpotqa}
Zhilin Yang, Peng Qi, Saizheng Zhang, Yoshua Bengio, William Cohen, Ruslan Salakhutdinov, and Christopher~D. Manning. 2018.
\newblock \href {https://aclanthology.org/D18-1259} {{H}otpot{QA}: A dataset for diverse, explainable multi-hop question answering}.
\newblock In \emph{Proceedings of the 2018 Conference on Empirical Methods in Natural Language Processing}, pages 2369--2380, Brussels, Belgium. Association for Computational Linguistics.

\bibitem[{Yao et~al.(2024)Yao, Qi, Pan, Cao, Hu, Liu, Hou, and Li}]{yao2024seakr}
Zijun Yao, Weijian Qi, Liangming Pan, Shulin Cao, Linmei Hu, Weichuan Liu, Lei Hou, and Juanzi Li. 2024.
\newblock Seakr: Self-aware knowledge retrieval for adaptive retrieval augmented generation.
\newblock \emph{arXiv preprint arXiv:2406.19215}.

\bibitem[{Yu et~al.(2024)Yu, Ping, Liu, Wang, You, Zhang, Shoeybi, and Catanzaro}]{yu2024rankrag}
Yue Yu, Wei Ping, Zihan Liu, Boxin Wang, Jiaxuan You, Chao Zhang, Mohammad Shoeybi, and Bryan Catanzaro. 2024.
\newblock \href {https://openreview.net/forum?id=S1fc92uemC} {Rank{RAG}: Unifying context ranking with retrieval-augmented generation in {LLM}s}.
\newblock In \emph{The Thirty-eighth Annual Conference on Neural Information Processing Systems}.

\bibitem[{Zhang et~al.(2025)Zhang, Liu, Hu, Niu, Wu, and Chen}]{zhang2025query}
Jiarui Zhang, Xiangyu Liu, Yong Hu, Chaoyue Niu, Fan Wu, and Guihai Chen. 2025.
\newblock Query routing for retrieval-augmented language models.
\newblock \emph{arXiv preprint arXiv:2505.23052}.

\bibitem[{Zhang et~al.(2024)Zhang, Patil, Jain, Shen, Zaharia, Stoica, and Gonzalez}]{zhang2024raft}
Tianjun Zhang, Shishir~G Patil, Naman Jain, Sheng Shen, Matei Zaharia, Ion Stoica, and Joseph~E. Gonzalez. 2024.
\newblock \href {https://openreview.net/forum?id=rzQGHXNReU} {{RAFT}: Adapting language model to domain specific {RAG}}.
\newblock In \emph{First Conference on Language Modeling}.

\bibitem[{Zhuang et~al.(2025)Zhuang, Ma, Koopman, Lin, and Zuccon}]{zhuang2025rank}
Shengyao Zhuang, Xueguang Ma, Bevan Koopman, Jimmy Lin, and Guido Zuccon. 2025.
\newblock Rank-r1: Enhancing reasoning in llm-based document rerankers via reinforcement learning.
\newblock \emph{arXiv preprint arXiv:2503.06034}.

\end{thebibliography}

\onecolumn
\appendix

\section{Prompt Template}
\label{sec:prompt_template}
The prompt template for different tasks are listed as follows.
\begin{figure*}[htbp]
\centering
\begin{tcolorbox}[
    colback=gray!15,
    colframe=gray!75,
    fonttitle=\large\bfseries\sffamily\color{white},
    coltitle=white,
    bottomrule=0pt,
    toprule=0pt,
    leftrule=0pt,
    rightrule=0pt,
    rounded corners,
]
\textbf{System Prompt:} You are a useful assistant. 
I will provide one question, 
several pieces of passages (which may be related or unrelated to the question). Please answer the question by selecting from one of the choice listed below. Please answer
with the capitalized alphabet only, without adding any extra phrase or period.  

\medskip
\textbf{User Prompt:} Passages: $\mathcal{P}_{q}$\\
Question: $q$\\

\medskip
\textbf{Assistant Prompt:} \{answer $a$\}
\end{tcolorbox}
\caption{Prompt for answer generation on multi-choice questions (e.g. MMLU, MMLU-Pro).}
\label{fig:hot_2wiki_rationale}
\end{figure*}

\begin{figure*}[htbp]
\centering
\begin{tcolorbox}[
    colback=gray!15,
    colframe=gray!75,
    fonttitle=\large\bfseries\sffamily\color{white},
    coltitle=white,
    bottomrule=0pt,
    toprule=0pt,
    leftrule=0pt,
    rightrule=0pt,
    rounded corners,
]
\textbf{System Prompt:} You are a useful assistant. 
I will provide one question, 
several pieces of passages (which may be related or unrelated to the question). Please answer the question with a short span containing one or few keywords.

\medskip
\textbf{User Prompt:} Passages: $\mathcal{P}_{q}$\\
Question: $q$\\

\medskip
\textbf{Assistant Prompt:} \{answer $a$\}
\end{tcolorbox}
\caption{Prompt for answer generation on span-based questions (e.g. SciQ, SimpleQA).}
\label{fig:hot_2wiki_rationale}
\end{figure*}

\begin{figure*}[htbp]
\centering
\begin{tcolorbox}[
    colback=gray!15,
    colframe=gray!75,
    fonttitle=\large\bfseries\sffamily\color{white},
    coltitle=white,
    bottomrule=0pt,
    toprule=0pt,
    leftrule=0pt,
    rightrule=0pt,
    rounded corners,
]
\textbf{System Prompt:} You are a useful assistant. 

\medskip
\textbf{User Prompt:} Here are a list of available external corpora:

pubmed: A high-quality biomedical corpus combining PubMed abstracts, focused on formal clinical language.

wikipedia: A cleaned and curated version of English Wikipedia containing factual and encyclopedic content across diverse topics.

c4: A filtered subset of the Colossal Clean Crawled Corpus, representing diverse high-quality web documents in English with broad topical coverage.

pes2o: The peS2o dataset is a collection of approximately 40 million open-access academic papers.

github: A curated collection of open-source GitHub repositories, featuring code, documentation, and  discussions centered on software engineering.

math: A dataset focused on mathematical reasoning and problem solving, containing questions, proofs, and solutions.

stackexchange: A cleaned snapshot of Stack Exchange QA threads, spanning technical, academic, and lifestyle topics.

book: A collection of long-form literary and non-fiction texts from public domain and licensed sources (e.g., PG19).

arxiv: A filtered set of LaTeX-based academic papers from arXiv, covering STEM domains with a focus on  technical writing.

commoncrawl: A nonprofit organization that provides a free, open repository of web crawl data to support research, analysis.

no: No retrieval. The model have enough knowledge to answer the question.

Given the question: \{question\}, identify the most appropriate external corpus to retrieve relevant information for answering it. Your response must be one of the source names listed above. If no external source is necessary, respond with no. 
Please directly output your predicted source with <source> and </source> tags.
\end{tcolorbox}
\caption{Prompt for question routing without chain-of-thought prompting.}
\label{fig:hot_2wiki_rationale}
\end{figure*}

\begin{figure*}[htbp]
\centering
\begin{tcolorbox}[
    colback=gray!15,
    colframe=gray!75,
    fonttitle=\large\bfseries\sffamily\color{white},
    coltitle=white,
    bottomrule=0pt,
    toprule=0pt,
    leftrule=0pt,
    rightrule=0pt,
    rounded corners,
]
\textbf{System Prompt:} You are a useful assistant. 

\medskip
\textbf{User Prompt:} Here are a list of available external corpora:

pubmed: A high-quality biomedical corpus combining PubMed abstracts, focused on formal clinical language.

wikipedia: A cleaned and curated version of English Wikipedia containing factual and encyclopedic content across diverse topics.

c4: A filtered subset of the Colossal Clean Crawled Corpus, representing diverse high-quality web documents in English with broad topical coverage.

pes2o: The peS2o dataset is a collection of approximately 40 million open-access academic papers.

github: A curated collection of open-source GitHub repositories, featuring code, documentation, and  discussions centered on software engineering.

math: A dataset focused on mathematical reasoning and problem solving, containing questions, proofs, and solutions.

stackexchange: A cleaned snapshot of Stack Exchange QA threads, spanning technical, academic, and lifestyle topics.

book: A collection of long-form literary and non-fiction texts from public domain and licensed sources (e.g., PG19).

arxiv: A filtered set of LaTeX-based academic papers from arXiv, covering STEM domains with a focus on  technical writing.

commoncrawl: A nonprofit organization that provides a free, open repository of web crawl data to support research, analysis.

no: No retrieval. The model have enough knowledge to answer the question.

Given the question: \{question\}, identify the most appropriate external corpus to retrieve relevant information for answering it. Your response must be one of the source names listed above. If no external source is necessary, respond with no. 
Please concise reasoning before output the final source. Please wrap your predicted source with <source> and </source> tags.

\end{tcolorbox}
\caption{Prompt for question routing with chain-of-thought prompting~\citep{wei2022chain}.}
\label{fig:hot_2wiki_rationale}
\end{figure*}

\newpage

\clearpage
\section{Detailed Per-task Performance}
\label{sec:task_performance}
The performance of different backbones on these six benchmarks with different knowledge is shown in the  Table \ref{tab:simpleqa} - \ref{table:csbench}. The results after adding the reranking stage is shown in Table \ref{tab:simpleqa_rerank} - \ref{tab:csbench_rerank}.

\begin{table*}[htbp]
\centering
\caption{Zero-shot Accuracy across different retrieval sources on the SimpleQA task. \vspace{-1ex}}
\label{tab:simpleqa}
\resizebox{\linewidth}{!}{
\begin{tabular}{llcccccccccccc}
\toprule
Dataset & Model & plain & Pubmed & Wiki & Pes2o & C4 & Github & Math & SE & Book & Arxiv & CC & All \\
\midrule
Simpleqa & Llama-3.2-3B & 0.034 & 0.014 & 0.251 & 0.020 & 0.054 & 0.015 & 0.023 & 0.019 & 0.031 & 0.015 & 0.095 & 0.296 \\
Simpleqa & Llama-3.1-8B & 0.009 & 0.011 & 0.286 & 0.025 & 0.079 & 0.021 & 0.031 & 0.033 & 0.042 & 0.018 & 0.120 & 0.299 \\
Simpleqa & Qwen3-4B     & 0.068 & 0.013 & 0.281 & 0.023 & 0.075 & 0.020 & 0.026 & 0.021 & 0.035 & 0.017 & 0.106 & 0.339 \\
Simpleqa & Qwen3-8B     & 0.079 & 0.011 & 0.293 & 0.027 & 0.079 & 0.025 & 0.032 & 0.026 & 0.042 & 0.019 & 0.121 & 0.345 \\
Simpleqa & Qwen3-32B    & 0.105 & 0.015 & 0.296 & 0.030 & 0.082 & 0.024 & 0.032 & 0.025 & 0.041 & 0.019 & 0.121 & 0.347 \\
Simpleqa & GPT-4o-mini  & 0.144 & 0.079 & 0.344 & 0.093 & 0.140 & 0.093 & 0.105 & 0.100 & 0.108 & 0.096 & 0.169 & 0.404 \\
Simpleqa & GPT-4o       & 0.343 & 0.258 & 0.425 & 0.253 & 0.275 & 0.249 & 0.262 & 0.248 & 0.273 & 0.258 & 0.288 & 0.463 \\
\bottomrule
\end{tabular}
}
\end{table*}

\begin{table*}[htbp]
\centering
\caption{Zero-shot accuracy of ARC-Challenge for different models and retrieval sources.\vspace{-1ex}}
\resizebox{\linewidth}{!}{
\begin{tabular}{llcccccccccccc}
\toprule
Dataset & Model & plain & Pubmed & Wiki & Pes2o & C4 & Github & Math & SE & Book & Arxiv & CC & All \\
\midrule
arc\_c & Llama-3.2-3B   & 0.626 & 0.572 & 0.629 & 0.579 & 0.639 & 0.583 & 0.627 & 0.564 & 0.639 & 0.570 & 0.637 & 0.650 \\
arc\_c & Llama-3.1-8B   & 0.693 & 0.682 & 0.718 & 0.720 & 0.724 & 0.687 & 0.724 & 0.652 & 0.742 & 0.717 & 0.738 & 0.736 \\
arc\_c & Qwen3-4B       & 0.853 & 0.829 & 0.823 & 0.827 & 0.846 & 0.819 & 0.849 & 0.814 & 0.849 & 0.817 & 0.852 & 0.850 \\
arc\_c & Qwen3-8B       & 0.890 & 0.851 & 0.853 & 0.844 & 0.864 & 0.863 & 0.866 & 0.846 & 0.874 & 0.846 & 0.868 & 0.860 \\
arc\_c & Qwen3-32B      & 0.928 & 0.905 & 0.906 & 0.904 & 0.919 & 0.915 & 0.916 & 0.903 & 0.917 & 0.904 & 0.911 & 0.903 \\
arc\_c & GPT-4o-mini    & 0.904 & 0.904 & 0.901 & 0.902 & 0.899 & 0.899 & 0.907 & 0.898 & 0.899 & 0.899 & 0.901 & 0.899 \\
arc\_c & GPT-4o         & 0.942 & 0.943 & 0.944 & 0.937 & 0.941 & 0.940 & 0.945 & 0.943 & 0.944 & 0.942 & 0.939 & 0.940 \\
\bottomrule
\end{tabular}
}
\end{table*}

\begin{table*}[htbp]
\centering
\caption{Zero-shot accuracy on SciQ for different models and retrieval sources.\vspace{-1ex}}
\resizebox{\linewidth}{!}{
\begin{tabular}{llcccccccccccc}
\toprule
Dataset & Model & plain & Pubmed & Wiki & Pes2o & C4 & Github & Math & SE & Book & Arxiv & CC & All \\
\midrule
sciq\_test & Llama-3.2-3B   & 0.465 & 0.364 & 0.515 & 0.388 & 0.560 & 0.338 & 0.507 & 0.362 & 0.517 & 0.321 & 0.580 & 0.598 \\
sciq\_test & Llama-3.1-8B   & 0.423 & 0.425 & 0.545 & 0.461 & 0.589 & 0.440 & 0.571 & 0.451 & 0.583 & 0.386 & 0.602 & 0.645 \\
sciq\_test & Qwen3-4B       & 0.530 & 0.451 & 0.565 & 0.432 & 0.610 & 0.491 & 0.561 & 0.476 & 0.580 & 0.470 & 0.591 & 0.641 \\
sciq\_test & Qwen3-8B       & 0.631 & 0.480 & 0.565 & 0.472 & 0.612 & 0.514 & 0.579 & 0.507 & 0.594 & 0.458 & 0.624 & 0.652 \\
sciq\_test & Qwen3-32B      & 0.705 & 0.494 & 0.621 & 0.512 & 0.641 & 0.500 & 0.599 & 0.507 & 0.622 & 0.471 & 0.658 & 0.660 \\
sciq\_test & GPT-4o-mini    & 0.690 & 0.562 & 0.586 & 0.525 & 0.636 & 0.556 & 0.586 & 0.574 & 0.613 & 0.537 & 0.608 & 0.634 \\
sciq\_test & GPT-4o         & 0.720 & 0.615 & 0.641 & 0.587 & 0.686 & 0.619 & 0.653 & 0.638 & 0.660 & 0.632 & 0.650 & 0.690 \\
\bottomrule
\end{tabular}}
\end{table*}

\begin{table*}[htbp]
\centering
\caption{Zero-shot accuracy on MMLU for different models and retrieval sources.\vspace{-1ex}}
\resizebox{\linewidth}{!}{
\begin{tabular}{llcccccccccccc}
\toprule
Dataset & Model & plain & Pubmed & Wiki & Pes2o & C4 & Github & Math & SE & Book & Arxiv & CC & All \\
\midrule
mmlu & Llama-3.2-3B   & 0.481 & 0.480 & 0.534 & 0.511 & 0.525 & 0.477 & 0.499 & 0.468 & 0.526 & 0.471 & 0.532 & 0.552 \\
mmlu & Llama-3.1-8B   & 0.530 & 0.528 & 0.576 & 0.574 & 0.580 & 0.547 & 0.565 & 0.521 & 0.590 & 0.554 & 0.588 & 0.606 \\
mmlu & Qwen3-4B       & 0.660 & 0.635 & 0.667 & 0.650 & 0.663 & 0.633 & 0.647 & 0.635 & 0.663 & 0.627 & 0.665 & 0.689 \\
mmlu & Qwen3-8B       & 0.689 & 0.668 & 0.688 & 0.679 & 0.689 & 0.670 & 0.686 & 0.666 & 0.689 & 0.658 & 0.697 & 0.713 \\
mmlu & Qwen3-32B      & 0.778 & 0.736 & 0.767 & 0.759 & 0.767 & 0.744 & 0.764 & 0.745 & 0.768 & 0.742 & 0.773 & 0.783 \\
mmlu & GPT-4o-mini    & 0.746 & 0.726 & 0.728 & 0.727 & 0.731 & 0.719 & 0.733 & 0.720 & 0.730 & 0.716 & 0.732 & 0.741 \\
mmlu & GPT-4o         & 0.833 & 0.821 & 0.823 & 0.820 & 0.823 & 0.824 & 0.824 & 0.823 & 0.823 & 0.823 & 0.826 & 0.828 \\
\bottomrule
\end{tabular}
}
\end{table*}

\begin{table*}[htbp]
\centering
\caption{Zero-shot accuracy on MMLU-Pro across different models and retrieval sources.\vspace{-1ex}}
\resizebox{\linewidth}{!}{
\begin{tabular}{llcccccccccccc}
\toprule
Dataset & Model & plain & Pubmed & Wiki & Pes2o & C4 & Github & Math & SE & Book & Arxiv & CC & All \\
\midrule
mmlu\_pro & Llama-3.2-3B   & 0.261 & 0.254 & 0.297 & 0.273 & 0.265 & 0.236 & 0.248 & 0.239 & 0.261 & 0.244 & 0.269 & 0.293 \\
mmlu\_pro & Llama-3.1-8B   & 0.301 & 0.308 & 0.356 & 0.344 & 0.352 & 0.325 & 0.350 & 0.317 & 0.363 & 0.324 & 0.365 & 0.389 \\
mmlu\_pro & Qwen3-4B       & 0.411 & 0.399 & 0.428 & 0.411 & 0.423 & 0.388 & 0.413 & 0.397 & 0.426 & 0.384 & 0.429 & 0.452 \\
mmlu\_pro & Qwen3-8B       & 0.446 & 0.428 & 0.450 & 0.438 & 0.451 & 0.427 & 0.458 & 0.435 & 0.456 & 0.420 & 0.461 & 0.482 \\
mmlu\_pro & Qwen3-32B      & 0.540 & 0.496 & 0.533 & 0.518 & 0.525 & 0.517 & 0.532 & 0.523 & 0.534 & 0.513 & 0.540 & 0.552 \\
mmlu\_pro & GPT-4o-mini    & 0.438 & 0.412 & 0.430 & 0.421 & 0.424 & 0.408 & 0.430 & 0.408 & 0.432 & 0.402 & 0.434 & 0.448 \\
mmlu\_pro & GPT-4o         & 0.554 & 0.544 & 0.547 & 0.547 & 0.548 & 0.548 & 0.547 & 0.546 & 0.552 & 0.547 & 0.549 & 0.556 \\
\bottomrule
\end{tabular}
}
\end{table*}

\begin{table*}[htbp]
\centering
\caption{Zero-shot accuracy on CSBench across different models and retrieval sources.}
\label{table:csbench}
\resizebox{\linewidth}{!}{
\begin{tabular}{llcccccccccccc}
\toprule
Dataset & Model & plain & Pubmed & Wiki & Pes2o & C4 & Github & Math & SE & Book & Arxiv & CC & All \\
\midrule
csbench & Llama-3.2-3B   & 0.440 & 0.414 & 0.485 & 0.460 & 0.474 & 0.466 & 0.480 & 0.474 & 0.478 & 0.460 & 0.477 & 0.489 \\
csbench & Llama-3.1-8B   & 0.472 & 0.435 & 0.528 & 0.506 & 0.529 & 0.514 & 0.536 & 0.507 & 0.527 & 0.500 & 0.537 & 0.542 \\
csbench & Qwen3-4B       & 0.653 & 0.576 & 0.639 & 0.629 & 0.634 & 0.597 & 0.642 & 0.629 & 0.630 & 0.607 & 0.641 & 0.661 \\
csbench & Qwen3-8B       & 0.698 & 0.602 & 0.649 & 0.638 & 0.661 & 0.642 & 0.667 & 0.645 & 0.652 & 0.631 & 0.663 & 0.673 \\
csbench & Qwen3-32B      & 0.760 & 0.676 & 0.719 & 0.706 & 0.723 & 0.715 & 0.715 & 0.718 & 0.721 & 0.700 & 0.721 & 0.722 \\
csbench & GPT-4o-mini    & 0.691 & 0.657 & 0.687 & 0.678 & 0.680 & 0.682 & 0.685 & 0.685 & 0.676 & 0.686 & 0.686 & 0.682 \\
csbench & GPT-4o         & 0.749 & 0.737 & 0.737 & 0.740 & 0.748 & 0.742 & 0.741 & 0.742 & 0.743 & 0.747 & 0.748 & 0.757 \\
\bottomrule
\end{tabular}
}
\end{table*}

\begin{table*}[htbp]
\centering
\caption{Zero-shot accuracy on SimpleQA with reranked retrieval sources.}
\label{tab:simpleqa_rerank}
\resizebox{\linewidth}{!}{
\begin{tabular}{llccccccccccc}
\toprule
Dataset & Model & Pubmed & Wiki & Pes2o & C4 & Github & Math & SE & Book & Arxiv & CC & All \\
\midrule
simpleqa & Llama-3.2-3B   & 0.013 & 0.273 & 0.022 & 0.066 & 0.015 & 0.026 & 0.017 & 0.031 & 0.014 & 0.098 & 0.320 \\
simpleqa & Llama-3.1-8B   & 0.035 & 0.290 & 0.048 & 0.088 & 0.043 & 0.047 & 0.041 & 0.052 & 0.037 & 0.122 & 0.309 \\
simpleqa & Qwen3-4B       & 0.012 & 0.306 & 0.025 & 0.082 & 0.021 & 0.027 & 0.021 & 0.035 & 0.017 & 0.120 & 0.364 \\
simpleqa & Qwen3-8B       & 0.012 & 0.312 & 0.030 & 0.091 & 0.019 & 0.030 & 0.025 & 0.044 & 0.019 & 0.134 & 0.382 \\
simpleqa & Qwen3-32B      & 0.015 & 0.313 & 0.030 & 0.090 & 0.023 & 0.034 & 0.026 & 0.044 & 0.020 & 0.133 & 0.369 \\
simpleqa & GPT-4o-mini    & 0.078 & 0.354 & 0.096 & 0.143 & 0.088 & 0.103 & 0.093 & 0.107 & 0.098 & 0.175 & 0.417 \\
simpleqa & GPT-4o         & 0.260 & 0.420 & 0.252 & 0.261 & 0.250 & 0.262 & 0.259 & 0.253 & 0.265 & 0.285 & 0.477 \\
\bottomrule
\end{tabular}
}
\end{table*}

\begin{table*}[htbp]
\centering
\caption{Zero-shot accuracy on ARC-Challenge with reranked retrieval sources.}
\resizebox{\linewidth}{!}{
\begin{tabular}{llccccccccccc}
\toprule
Dataset & Model & Pubmed & Wiki & Pes2o & C4 & Github & Math & SE & Book & Arxiv & CC & All \\
\midrule
arc\_c & Llama-3.2-3B   & 0.586 & 0.642 & 0.600 & 0.641 & 0.570 & 0.612 & 0.568 & 0.621 & 0.564 & 0.639 & 0.658 \\
arc\_c & Llama-3.1-8B   & 0.692 & 0.718 & 0.709 & 0.730 & 0.676 & 0.715 & 0.664 & 0.729 & 0.691 & 0.732 & 0.755 \\
arc\_c & Qwen3-4B       & 0.817 & 0.824 & 0.829 & 0.842 & 0.816 & 0.848 & 0.811 & 0.837 & 0.812 & 0.848 & 0.848 \\
arc\_c & Qwen3-8B       & 0.851 & 0.856 & 0.846 & 0.864 & 0.852 & 0.878 & 0.846 & 0.868 & 0.835 & 0.859 & 0.860 \\
arc\_c & Qwen3-32B      & 0.902 & 0.907 & 0.897 & 0.916 & 0.911 & 0.917 & 0.905 & 0.911 & 0.888 & 0.913 & 0.914 \\
arc\_c & GPT-4o-mini    & 0.903 & 0.904 & 0.900 & 0.901 & 0.903 & 0.903 & 0.902 & 0.904 & 0.899 & 0.904 & 0.902 \\
arc\_c & GPT-4o         & 0.942 & 0.933 & 0.936 & 0.939 & 0.943 & 0.943 & 0.945 & 0.945 & 0.938 & 0.942 & 0.944 \\
\bottomrule
\end{tabular}
}
\end{table*}

\begin{table*}[h]
\centering
\caption{Zero-shot accuracy on SciQ with reranked retrieval sources.}
\resizebox{\linewidth}{!}{
\begin{tabular}{llccccccccccc}
\toprule
Dataset & Model & Pubmed & Wiki & Pes2o & C4 & Github & Math & SE & Book & Arxiv & CC & All \\
\midrule
sciq\_test & Llama-3.2-3B   & 0.363 & 0.531 & 0.420 & 0.577 & 0.342 & 0.514 & 0.372 & 0.505 & 0.315 & 0.575 & 0.654 \\
sciq\_test & Llama-3.1-8B   & 0.418 & 0.543 & 0.450 & 0.625 & 0.425 & 0.581 & 0.436 & 0.581 & 0.364 & 0.616 & 0.678 \\
sciq\_test & Qwen3-4B       & 0.449 & 0.566 & 0.480 & 0.629 & 0.477 & 0.576 & 0.470 & 0.578 & 0.428 & 0.627 & 0.669 \\
sciq\_test & Qwen3-8B       & 0.473 & 0.551 & 0.497 & 0.623 & 0.487 & 0.593 & 0.502 & 0.579 & 0.450 & 0.620 & 0.689 \\
sciq\_test & Qwen3-32B      & 0.482 & 0.602 & 0.536 & 0.638 & 0.507 & 0.630 & 0.512 & 0.626 & 0.460 & 0.655 & 0.698 \\
sciq\_test & GPT-4o-mini    & 0.551 & 0.589 & 0.515 & 0.620 & 0.557 & 0.593 & 0.567 & 0.593 & 0.523 & 0.628 & 0.657 \\
sciq\_test & GPT-4o         & 0.619 & 0.625 & 0.587 & 0.672 & 0.629 & 0.656 & 0.628 & 0.651 & 0.624 & 0.665 & 0.695 \\
\bottomrule
\end{tabular}
}
\end{table*}

\begin{table*}[h]
\centering
\caption{Zero-shot accuracy on MMLU with reranked retrieval sources.}
\resizebox{\linewidth}{!}{
\begin{tabular}{llccccccccccc}
\toprule
Dataset & Model & Pubmed & Wiki & Pes2o & C4 & Github & Math & SE & Book & Arxiv & CC & All \\
\midrule
mmlu & Llama-3.2-3B   & 0.480 & 0.535 & 0.505 & 0.523 & 0.469 & 0.499 & 0.468 & 0.523 & 0.459 & 0.528 & 0.559 \\
mmlu & Llama-3.1-8B   & 0.526 & 0.582 & 0.567 & 0.576 & 0.535 & 0.557 & 0.523 & 0.577 & 0.540 & 0.579 & 0.612 \\
mmlu & Qwen3-4B       & 0.634 & 0.658 & 0.648 & 0.665 & 0.631 & 0.646 & 0.627 & 0.661 & 0.623 & 0.670 & 0.685 \\
mmlu & Qwen3-8B       & 0.658 & 0.685 & 0.673 & 0.689 & 0.665 & 0.679 & 0.663 & 0.686 & 0.652 & 0.697 & 0.710 \\
mmlu & Qwen3-32B      & 0.733 & 0.763 & 0.755 & 0.759 & 0.743 & 0.759 & 0.743 & 0.762 & 0.738 & 0.768 & 0.780 \\
mmlu & GPT-4o-mini    & 0.727 & 0.733 & 0.728 & 0.729 & 0.720 & 0.730 & 0.724 & 0.730 & 0.717 & 0.734 & 0.741 \\
mmlu & GPT-4o         & 0.822 & 0.820 & 0.820 & 0.823 & 0.822 & 0.826 & 0.823 & 0.821 & 0.821 & 0.824 & 0.827 \\
\bottomrule
\end{tabular}
}
\end{table*}

\begin{table*}[h]
\centering
\caption{Zero-shot accuracy on MMLU-Pro with reranked retrieval sources.}
\resizebox{\linewidth}{!}{
\begin{tabular}{llccccccccccc}
\toprule
Dataset & Model & Pubmed & Wiki & Pes2o & C4 & Github & Math & SE & Book & Arxiv & CC & All \\
\midrule
mmlu\_pro & Llama-3.2-3B   & 0.260 & 0.300 & 0.279 & 0.279 & 0.245 & 0.261 & 0.251 & 0.273 & 0.244 & 0.280 & 0.297 \\
mmlu\_pro & Llama-3.1-8B   & 0.310 & 0.362 & 0.337 & 0.351 & 0.318 & 0.345 & 0.315 & 0.358 & 0.316 & 0.361 & 0.395 \\
mmlu\_pro & Qwen3-4B       & 0.398 & 0.424 & 0.408 & 0.423 & 0.388 & 0.419 & 0.396 & 0.423 & 0.381 & 0.425 & 0.454 \\
mmlu\_pro & Qwen3-8B       & 0.420 & 0.449 & 0.435 & 0.451 & 0.424 & 0.454 & 0.431 & 0.457 & 0.418 & 0.458 & 0.485 \\
mmlu\_pro & Qwen3-32B      & 0.499 & 0.527 & 0.518 & 0.527 & 0.514 & 0.529 & 0.520 & 0.535 & 0.506 & 0.535 & 0.553 \\
mmlu\_pro & GPT-4o-mini    & 0.412 & 0.427 & 0.417 & 0.423 & 0.403 & 0.429 & 0.408 & 0.430 & 0.400 & 0.433 & 0.459 \\
mmlu\_pro & GPT-4o         & 0.544 & 0.547 & 0.544 & 0.546 & 0.551 & 0.552 & 0.548 & 0.551 & 0.546 & 0.549 & 0.558 \\
\bottomrule
\end{tabular}
}
\end{table*}

\begin{table*}[h]
\centering
\caption{Zero-shot accuracy on CSBench with reranked retrieval sources.}
\label{tab:csbench_rerank}
\resizebox{\linewidth}{!}{
\begin{tabular}{llccccccccccc}
\toprule
Dataset & Model & Pubmed & Wiki & Pes2o & C4 & Github & Math & SE & Book & Arxiv & CC & All \\
\midrule
csbench & Llama-3.2-3B   & 0.412 & 0.478 & 0.458 & 0.480 & 0.462 & 0.471 & 0.463 & 0.466 & 0.447 & 0.473 & 0.501 \\
csbench & Llama-3.1-8B   & 0.427 & 0.528 & 0.497 & 0.523 & 0.511 & 0.525 & 0.520 & 0.533 & 0.507 & 0.541 & 0.547 \\
csbench & Qwen3-4B       & 0.571 & 0.644 & 0.624 & 0.643 & 0.613 & 0.655 & 0.623 & 0.643 & 0.596 & 0.634 & 0.658 \\
csbench & Qwen3-8B       & 0.602 & 0.658 & 0.638 & 0.661 & 0.650 & 0.664 & 0.643 & 0.651 & 0.619 & 0.669 & 0.676 \\
csbench & Qwen3-32B      & 0.684 & 0.713 & 0.694 & 0.714 & 0.712 & 0.726 & 0.728 & 0.727 & 0.688 & 0.728 & 0.722 \\
csbench & GPT-4o-mini    & 0.668 & 0.683 & 0.679 & 0.679 & 0.680 & 0.681 & 0.691 & 0.690 & 0.682 & 0.686 & 0.691 \\
csbench & GPT-4o         & 0.733 & 0.744 & 0.736 & 0.750 & 0.737 & 0.744 & 0.739 & 0.746 & 0.741 & 0.748 & 0.746 \\
\bottomrule
\end{tabular}
}
\end{table*}

\end{document}